\documentclass[11pt]{article}

\usepackage[utf8]{inputenc} 
\usepackage[T1]{fontenc}    
\usepackage{lmodern}
\usepackage{url}            
\usepackage{booktabs}       
\usepackage{amsfonts}       
\usepackage{nicefrac}       
\usepackage{microtype}      
\usepackage{float}
\usepackage{enumerate}

\usepackage{caption}
\usepackage{mathrsfs} 
\usepackage{amsmath}
\usepackage{amssymb}
\usepackage{graphicx}
\usepackage{url}
\PassOptionsToPackage{x11names}{xcolor}
\usepackage{xcolor}
\PassOptionsToPackage{algo2e,ruled}{algorithm2e}
\usepackage{algorithm2e}
\usepackage{jmlrutils}
\usepackage{hyperref}
\usepackage[titletoc]{appendix}
\usepackage{authblk}

\usepackage{nameref}

\usepackage{authblk}

\usepackage{pdfpages} 

\usepackage[margin= 0.8in,top=12mm]{geometry}

\newcommand*{\titleAT}{\begingroup
  \newlength{\drop}
  \drop=0.05\textheight   
  \begin{center}
  \includegraphics[scale=0.4]{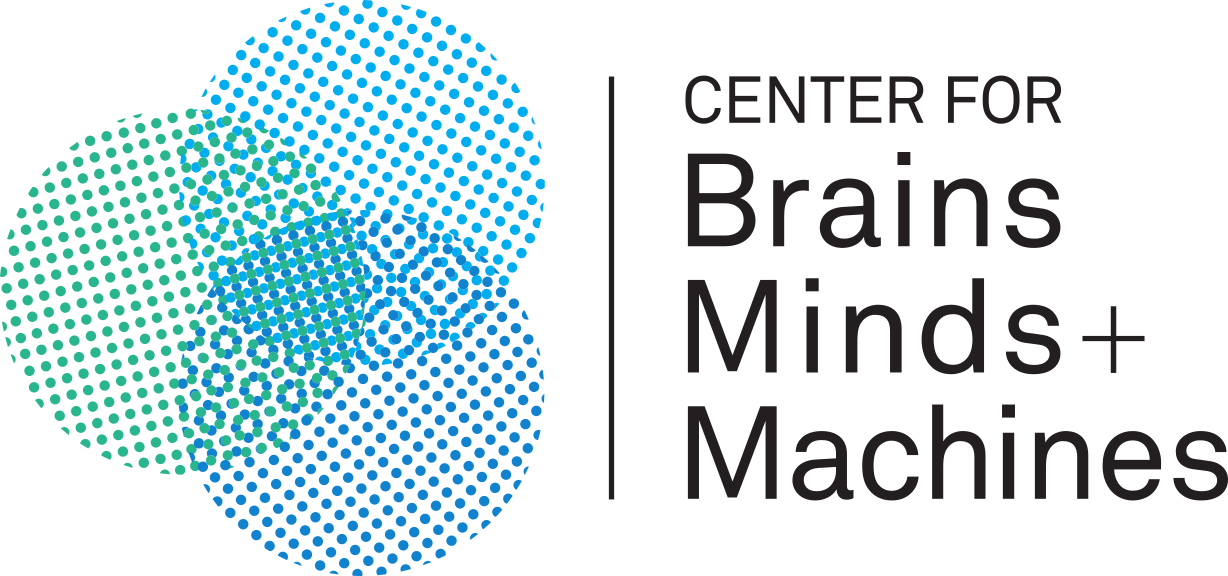} 
  \end{center} 
  \vspace{2pt}\vspace{-\baselineskip}

  \vspace{\drop}
  \textbf{\large{CBMM Memo No. \memonumber}}   \hfill    \textbf{\large{\memodate}} 

  \vspace{\drop}
  \begin{center}
    \textbf{\huge{\memotitle}}\\
    \vspace{0.2\drop}  
    \large{\memoauthors}
  \end{center}
  \vspace{\drop} 
  \textbf{\large{\noindent Abstract}:} {\memoabstract}

\vspace{\fill}     
  \rule{\textwidth}{0.4pt}\par
  \begin{minipage}{.15\linewidth}
    \includegraphics[scale=0.1]{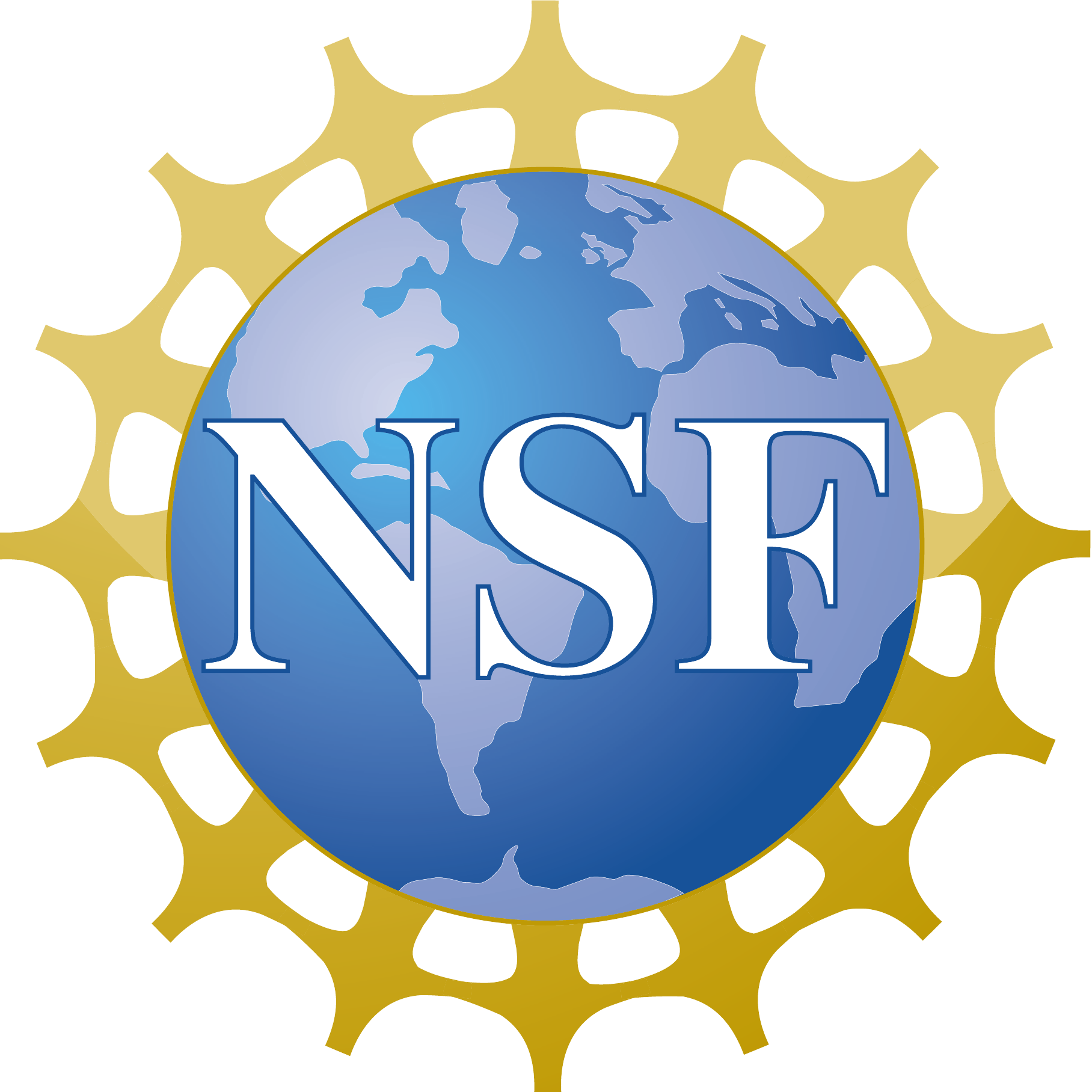}
  \end{minipage}
  \begin{minipage}{.84\linewidth}
    \textbf{\large{This work was supported by the Center for Brains, Minds and Machines (CBMM), funded by NSF STC award  CCF - 1231216.}}
  \end{minipage}
  \endgroup}

\begin{document}

\def\memonumber{91}        
\def\memodate{\today}
\def\memotitle {A Surprising Linear Relationship Predicts Test Performance in Deep Networks} 
\def\memoauthors{
Qianli Liao$^{1}$, Brando Miranda$^{1}$,  Andrzej Banburski$^{1}$,  Jack Hidary$^{2}$ and Tomaso Poggio$^{1}$ \\
$^1$Center for Brains, Minds, and Machines, MIT\\
$^2$Alphabet (Google) X    
}

\normalsize 
\def\memoabstract{
 \footnote{A previous version of this paper is \cite{liao2018classical}.}       
            Given two networks with the same training loss on a
            dataset, when would they have drastically different test
            losses and errors? Better understanding of this question
            of generalization may improve practical applications of
            deep networks. In this paper we show that with cross-entropy
            loss it is surprisingly simple to induce        
            significantly different generalization performances for     
            two networks that have the same architecture, the same
            meta parameters and the same training error: one can 
            either pretrain the networks with different levels of
            "corrupted" data or simply initialize the networks with
            weights of different Gaussian standard deviations. A
            corollary of recent theoretical results on overfitting
            shows that these effects are due to an intrinsic problem
            of measuring test performance with a cross-entropy/exponential-type      
            loss, which can be decomposed into two components both
            minimized by SGD --- one of which is not related to
            expected classification performance. However, if we factor
            out this component of the loss, a linear relationship
            emerges between training and test losses. Under this
            transformation, classical generalization bounds are
            surprisingly tight: the empirical/training loss is very
            close to the expected/test loss. Furthermore, the
            empirical relation between classification error and
            normalized cross-entropy loss seem to be approximately
            monotonic.    
  
} 
\titleAT

        \newpage
        \tableofcontents
        \newpage 

\section{Introduction}

Despite many successes of deep networks and a growing amount of
research, several fundamental questions remain unanswered,
among which is a key puzzle is about the apparent lack of generalization,
defined as convergence of the training performance to the test
performance with increasing size of the training set. How can a
network predict well without generalization?  What is the relationship
between training and test performances?

In this paper, we investigate the question of why two deep networks
with the same training loss have different testing performances. This
question is valuable in theory and in practice since training loss is
an important clue for choosing among different deep learning
architectures. It is therefore worth studying how much training loss
can tell us about generalization. In the rest of the paper, we will
use the term ``error'' to mean ``classification error'' and the term
``loss'' to mean ``cross-entropy loss'', the latter being the
objective minimized by stochastic gradient descent (SGD).
 
In addition to training error and loss, there are many factors (such
as choices of network architecture) that can affect generalization
performance and we cannot exhaustively study them in this
paper. Therefore, we restrict our models to have the same architecture
and training settings within each experiment. We tried different
architectures in different experiments and observed consistent
results.

\section{Observation: Networks with the Same Training Performance Show
  Different Test Performance}      

First we start with a common observation: even when two networks have
the same architecture, same optimization meta parameters and same
training loss, they usually have different test performances
(i.e. error and loss), presumably because the stochastic nature of the  
minimization process converge to different minima among the many
existing in the loss landscape \cite{Theory_II,Theory_III,
  Theory_IIIb}.

With standard settings the differences are usually small (though
significant, as shown later). We propose therefore two approaches to
magnify the effect:
\begin{itemize}
\item Initialize networks with different levels of ``random
  pretraining'': the network is pretrained for a specified number of
  epochs on ``corrupted'' training data --- the labels of a portion of
  the examples are swapped with each other in a random fashion.
\item Initialize the weights of the networks with different standard
  deviations of a diagonal Gaussian distribution. As it turns out, different
  standard deviations yield different test   performance.
\end{itemize}

\begin{figure}[H]\centering
	\includegraphics[width=0.75\textwidth]{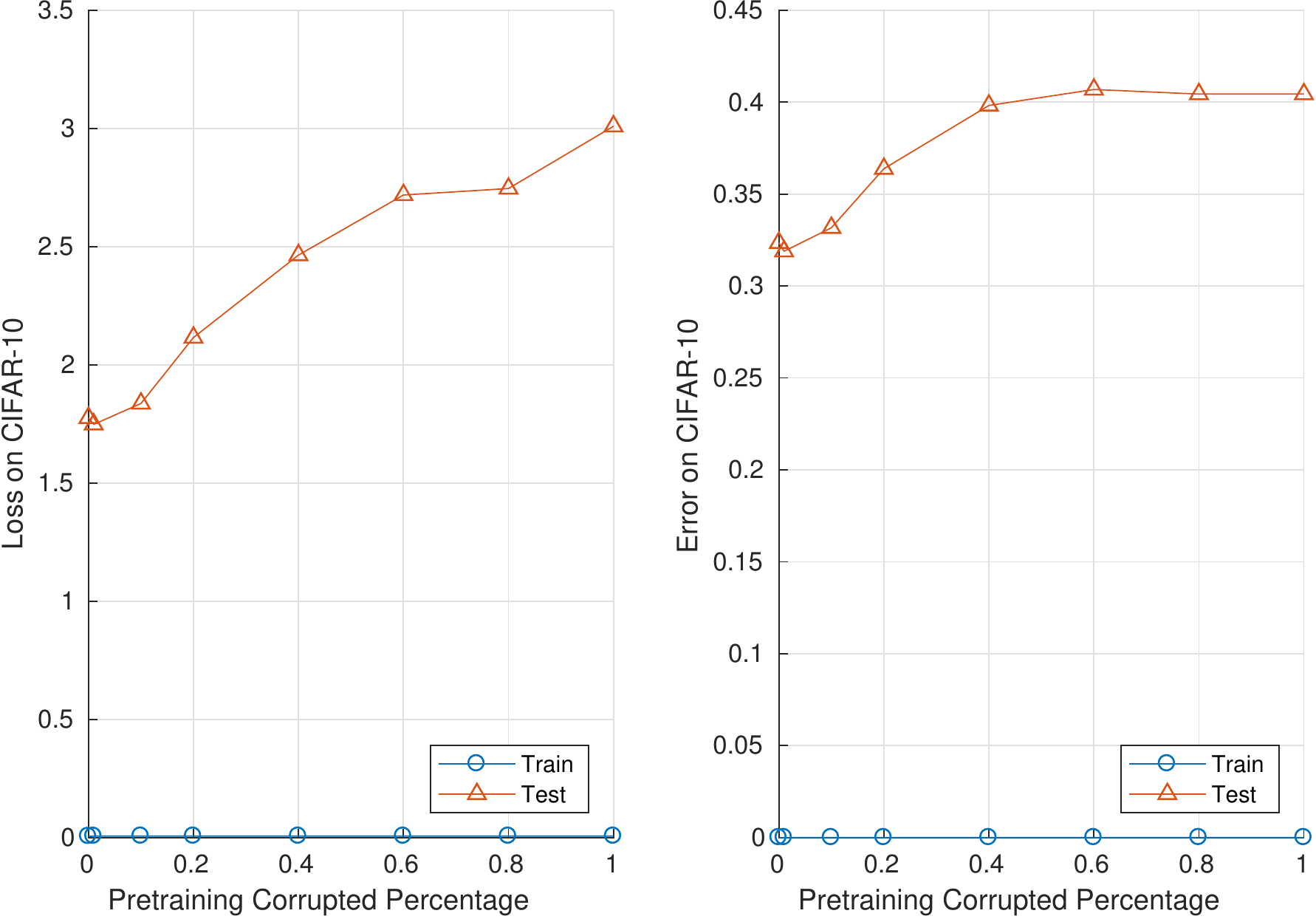}  
	\caption{\it Random Pretraining vs Generalization Performance
          on CIFAR-10: a 5-layer ConvNet (described in section
          \ref{5layers}) is pretrained on training data with partially
          ``corrupted'' labels for 30 epochs. It is then trained on
          normal data for 80 epochs. Among the network snapshots
          saved from all the epochs we pick a network that is closest
          to an arbitrarily (but low enough) chosen reference  training loss (0.006 here).
          The number on the x axis indicates the percentage of labels
          that are swapped randomly. As pretraining data gets
          increasingly ``corrupted'', the generalization performance
          of the resultant model becomes increasingly worse, even
          though they have similar training losses and the same zero
          classification error in training. Batch normalization (BN)
          is used. After training, the means and standard deviations
          of BN are ``absorbed'' into the network's weights and
          biases. No data augmentation is performed.  }
	\label{cifar10_RLNL_performance}
\end{figure}

\begin{figure}[H]\centering  
	\includegraphics[width=0.75\textwidth]{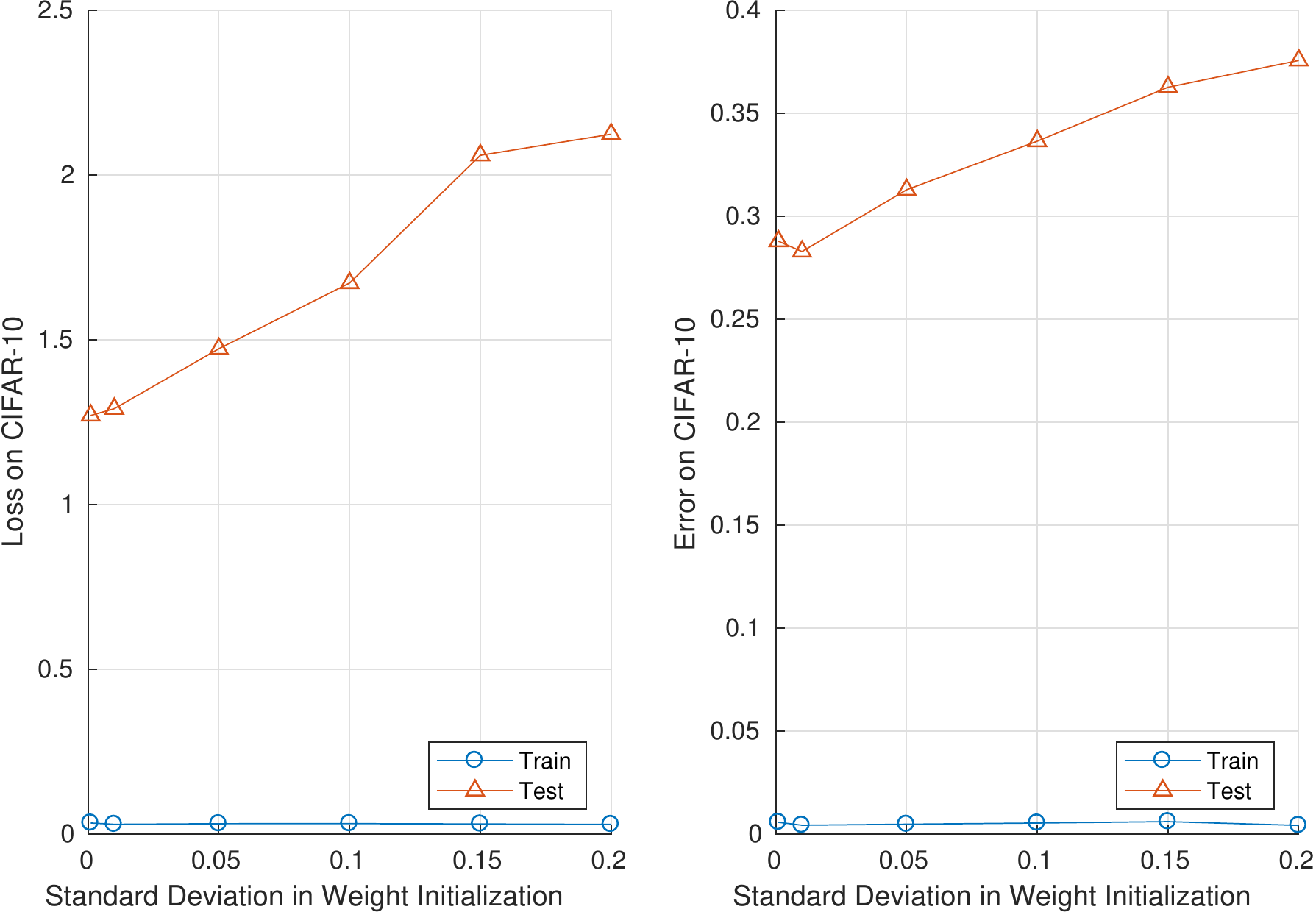}  
	\caption{\it Standard Deviation in weight initialization vs
          generalization performance on CIFAR-10: the network is
          initialized with weights of different standard
          deviations. The other settings are the same as in Figure
          \ref{cifar10_RLNL_performance}. As the norm of the initial
          weights becomes larger, the generalization performance of
          the resulting model is worse, even though all models have
          the same classification error in training.  }
	\label{cifar10_DMag_performance}
\end{figure}

We show the results of ``random pretraining'' with networks on
CIFAR-10 (Figure \ref{cifar10_RLNL_performance}) and CIFAR-100 (Figure
\ref{cifar100_RLNL_performance}) and initialization with different
standard deviations on CIFAR-10 (Figure
\ref{cifar10_DMag_performance}) and CIFAR-100 (Figure
\ref{cifar100_DMag_performance}).

\section{Theory: Measuring Generalization Correctly}

In the previous section, we observed that it is possible to obtain
networks that have the same training loss (and error) but very
different test performances. This indicates that training loss is not
a good proxy of test loss. In general, it is quite common to get zero
training error (and a very small training loss) when using
overparametrized networks. Therefore, comparing models based on
training statistics becomes very difficult. In an extreme
case of this situation, recent work \cite{zhang2016understanding}
shows that an overparametrized deep network can fit a randomly labeled
training set and of course fail to generalize at all. In some sense,
deep nets seem to ``fool'' training loss: extremely low training loss
can be achieved without any generalization.

Previous theoretical work\cite{Theory_IIIb} has provided an answer to
the above puzzle. In this paper, we show a corollary of the theory and
motivate it in a simple way showing why this ``super-fitting''
phenomenon happens and under which conditions generalization holds for deep networks.

\subsection{The Normalized Cross-Entropy Loss}  

\textbf{Notation:} We define (as in \cite{Theory_IIIb}) a deep network
with $K$ layers with the usual elementwise scalar activation functions
$\sigma(z):\quad \mathbf{R} \to \mathbf{R}$ as the set of functions
$f(W;x) = \sigma (W^K \sigma (W^{K-1} \cdots \sigma (W^1 x)))$, where
the input is $x \in \mathbf{R}^d$, the weights are given by the
matrices $W^k$, one per layer, with matching dimensions. We use the
symbol $W$ as a shorthand for the set of $W^k$ matrices
$k=1,\cdots,K$. For simplicity we consider here the case of binary
classification in which $f$ takes scalar values, implying that the
last layer matrix $W^K$ is $W^K \in \mathbf{R}^{1,K_l}$. There are no
biases apart form the input layer where the bias is instantiated by
one of the input dimensions being a constant. The activation function
in this paper is the ReLU activation.
        
\label{Predicting expected error}

Consider different zero minima of the empirical risk obtained with the
same network on the same training set. Can we predict their expected
error from empirical properties only? A natural way to approach the
problem of ranking two different minimizers of the empirical risk
starts with the ``positive homogeneity'' property of ReLU networks
(see \cite{Theory_IIIb}) :

\begin{equation}
f(W^1,\cdots,W_k;x)=\rho_1, \dotsm, \rho_K f(\tilde{W^1},\cdots,\tilde{W_k};x)
\end{equation}
\noindent where $W_k=\rho_k \tilde{W_k}$ and $||\tilde{W_k}||=1$. 

This property is valid for layerwise normalization under any 
norm. Note that $f(W^1,\cdots,W_k;x)$ and
$f(\tilde{W^1},\cdots,\tilde{W_k};x) $ have the same classification
performance on any given (test) set. It follows that different empirical 
minimizers should be compared in terms of their normalized form: the
$\rho$ factors affect an exponential type loss  -- driving it to zero by
increasing the $\rho$s to infinity -- but do not change the classification
performance which only depends on the sign of $y_n f(x_n)$. 
Consider the cross-entropy for two classes, given by
 
\begin{equation}
L = \sum_{n=1}^N \ell(f(x_n), yn)=\sum_{n=1}^N \ln (1+ e^{-y_n f(x_n)})= \sum_{n=1}^N \ln(1+e^{-y_n (\rho_1, \dotsm, \rho_K) f(\tilde{W^1},\cdots,\tilde{W_k};x_n) }).
\end{equation}

What is the right norm to be used to normalize
the capacity of deep networks? The ``right'' normalization should make
different minimizers equivalent from the point of view of
their intrinsic capacity -- for instance equivalent in terms of
their Rademacher complexity. The positive homogeneity property
implies that the correct norm for a multilayer network should be based
on a product of norms. Since different norms are equivalent in
$\mathbb{R}^n$ it is enough to use $L_2$.

The argument can be seen more directly looking at a typical
generalization bound (out of many) of the form
\cite{Bousquet2003}, where now $l$ is the bounded loss function $l(y,
f(W; x))$:

{\it With probability
  $\geq (1-\delta)$ $\forall g$

\begin{equation}
|\mathbf{E} (l) - \mathbf{E}_S(l)| \leq  c_1
\mathbb{C}_N(\mathbf{L}) + c_2 \sqrt \frac{\ln(\frac{1}{\delta})}{2N} 
\label{bound}
\end{equation} 
}
\noindent where $\mathbf{E} (l)$ is the expected loss,
$\mathbf{E}_S(l)$ is the empirical loss, $N$ is the size of the
training set, $\mathscr{C}_N(\mathbf{G})$ is an empirical complexity
of the class of functions $\mathbf{G}$ on the unit sphere (with
respect to each weight per layer) to which $l$ belongs and $c_1$ and
$c_2$ are constant depending on properties of the loss function. The
prototypical example of such a complexity measure is the empirical
Rademacher average, typically used for classification in a bound of
the form \ref{bound}. Another example involves covering numbers. We
call the bound ``tight'' if the right hand side -- that we call
``offset'' -- is small. Notice that a very small offset, assuming that
the bound holds, implies a linear relation with slope $1$ between
$\mathbf{E} (l)$ and $\mathbf{E}_S(l)$. Thus a very tight bound
implies linearity with slope $1$.  Many learning algorithms have the
closely related property of ``generalization'':
$\lim_{N \to \infty}|\mathbf{E} (l) - \mathbf{E}_S(l)| \to 0$ (a
notable exception is the nearest neighbor classifier), since in
general we hope that $\mathbb{C}_N(\mathbf{G}) $ is such to decrease
as $N$ increases.

We expect layerwise normalization to yield the same complexity for 
the different minimizers. The case of linear functions and binary
classification is an example. The Rademacher complexity of a set
$\mathbf{F})$ of linear functions $f(x) = w^T x$ can be bounded as
$\mathscr{R}_N(\mathbf{F}) \leq c \sqrt \frac{1}{n} X W$ where $X$ is an $L_p$ bound on
the vectors $x$ and $W$ is an $L_q$ bound on the vectors $w$ with
$L_p$ and $L_q$ being dual norms, that is
$\frac{1}{p}+\frac{1}{q}=1$. This includes $L_2$ for both $w$ and $x$
but also $L_1$ and $L_\infty$. Since different $L_p$ norms are
equivalent in finite dimensional spaces (in fact $||x||_p \ge ||x||_q$
when $p \leq q$, with opposite relations holding with appropriate
constants, e.g. $||x||_p \leq n^{1/p-1/q} ||x||_q$ for $x\in\mathbb{R}^n$), Equation \ref{bound} holds under normalization with
different $L_p$ norms for $w$ (in the case of linear networks). Notice
that the other term in the bound (the last term on the right-hand side
of Equation \ref{bound}) is the same for all networks trained on the
same dataset. 

The results of \cite{Theory_IIb} -- which, as mentioned earlier,  was 
the original motivation behind the arguments above and the experiments
below-- imply that the weight matrices at each layer converge to the 
minimum Frobenius norm for each minimizer.

\subsection{Bounds on the Cross-Entropy Loss imply Bounds on the
  Classification Error}

Running SGD on the cross-entropy loss in deep networks, as usually
done, is a typical approach in machine learning: minimize a convex and differentiable
{\it surrogate} of the $0-1$ loss function (in the case of binary classification).
The approach is based on the fact that  the logistic loss (for two
classes, cross-entropy becomes the logistic loss)  is an upper bound to
the binary classification error: thus minimization of the loss implies
minimization of the error. 
One way to formalize these upper bounds is to consider the excess
classification risk $R(f)-R^*$, where $R(f)$ is the classification
loss associated with $f$ and $R^*$ is the Bayes risk \cite{Bartlett03convexity}. Let us call
$R_{\ell}(f) = \mathbf{E} (\ell)$ the cross-entropy risk and $R_{\ell}^*$ the optimal cross
entropy risk.  Then the following bound holds in terms of the 
so-called $\psi$-transform of the logistic loss $\ell$:

\begin{equation}
  \psi(R(f) - R^*) \leq R_{\ell}(f) - R^*_{\ell})
\label{psitransformbound}
\end{equation} 

\noindent where the $\psi$ function for the exponential loss -- a good
proxy for the logistic loss -- can be computed analytically as
$\psi(x) = 1- \sqrt(1-x^2)$. Interestingly the shape of $\psi(x)$ for
$0 \leq x \leq 1$ is quite similar to the empirical dependence of classification
error on normalized cross-entropy loss in figures such as the bottom
right of Figure \ref{RLNLcifar10}.

Unfortunately lower bounds are not available. As a consequence, we
cannot prove that among two networks the one with  better normalized
training cross-entropy (which implies a similar loss at test)
guarantees a lower classification error at test. This difficulty is
not specific to deep networks or to our main result. It is common to
all optimization techniques using a surrogate loss function.

\section{Experiments: Normalization Leads to Surprising Tight Generalization}         
\label{expected loss is linear in normalized training loss}

In this section we discuss the experiment results after normalizing
the capacity of the networks, as discussed in the previous section.
What we observe is a linear relationship between the train loss and
the test loss, implying that the expected loss is very close to the
empirical loss.

\subsection{Tight Linearity}

Our observations are that the linear relationship between train loss
and test loss hold in a robust way under several conditions:

\begin{itemize}
\item {\it Independence from Initialization:} The linear relationship
  is independent of whether the initialization is via pretraining on
  randomly labeled natural images or whether it is via larger
  initialization, as shown by Figures \ref{RLNLcifar10} and
  \ref{SDcifar10}.  In particular, notice that the left of Figure
  \ref{testlossvstrainloss} and Figure \ref{only_NL_points} clearly
  shows a linear relation for minimizers obtained with the default
  initialization in Pytorch \cite{paszke2017automatic, EfficientBackPropLeCun}.  
  This point is important because it
  shows that the linear relationship is not an artifact of pretraining
  on random labels.
\item {\it Independence from Network Architecture:} The linear
  relationship of the test loss and train loss is independent from the
  network architectures we tried.  Figure \ref{not_same_train_loss},
  \ref{NL_vs_RL} show the linear relationship for a $3$ layer network
  without batch normalization while Figures \ref{RLNLcifar10},
  \ref{SDcifar10} show the linear relationship for a $5$ layer network with batch
  normalization on CIFAR10.  Additional evidence can be found in the
  appendix in Figures \ref{L1}, \ref{testlossvstrainloss} and
  \ref{RL_point}.
  
\item {\it Independence from Data Set:} Figures
  \ref{not_same_train_loss}, \ref{RLNLcifar10}, \ref{SDcifar10},
  \ref{NL_vs_RL}, \ref{L1}, \ref{testlossvstrainloss},
  \ref{only_NL_points}, \ref{RL_point} show the linear relationship on
  CIFAR10 while Figures \ref{RLNLcifar100} and \ref{SDcifar100} show
  the linear relationship on CIFAR100.

  \item {\it Norm independence:} Figures \ref{L1} show that the $L_p$ norm
  used for normalization does not matter -- as expected.   
  
\item {\it Normalization is independent of training loss:} Figure
  \ref{not_same_train_loss} shows that networks with different
  cross-entropy training losses (which are sufficiently large to
  guarantee zero classification error), once normalized, show
  the same linear relationship between train loss and test loss.
 
\end{itemize}
\subsection{Randomly Labeled Training Data}

Since the empirical complexity in Equation \ref{bound} does
not depend on the labels of the training set, Equation \ref{bound}
should also hold in predicting expected error when the training is on
randomly labeled data \cite{DBLP:journals/corr/ZhangBHRV16} (though
different stricter bounds may hold separately for each of the two
cases). The experiments in Figures \ref{RL_point}, \ref{RLNLcifar10}
and \ref{RLNLcifar100} show exactly this result. The new datapoint
(corresponding to the randomly trained network) is still a tight bound:
{\it the empirical loss is very similar to the test loss}. This is quite
{\it surprising}: the corresponding classification error is zero for
training and at chance level for testing!

\subsection{Higher Capacity leads to Higher Test Error}

\begin{figure}[H]
  \centering
  \includegraphics[width=1.0\textwidth]{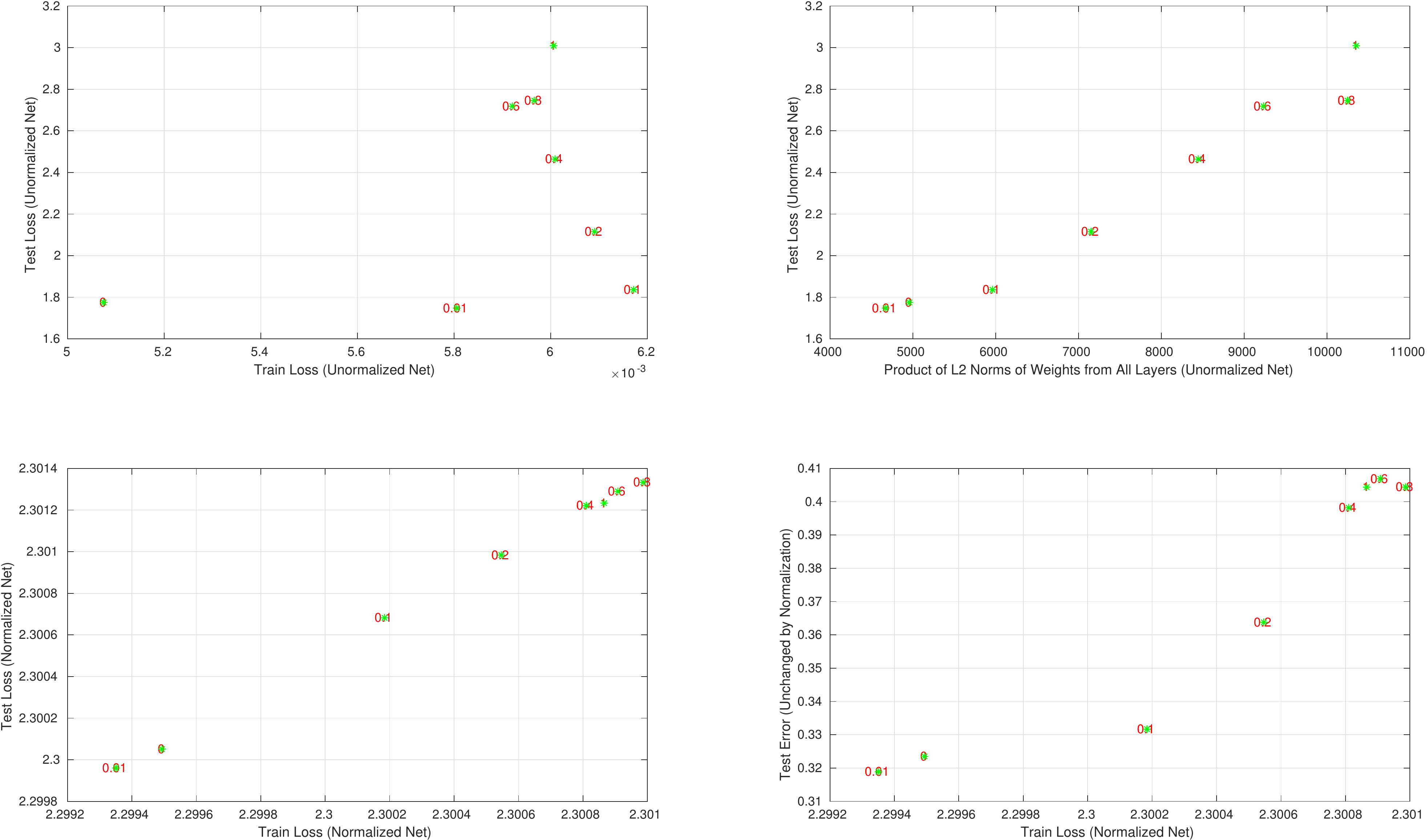}     
  \caption{\it Random pretraining experiment with batch
    normalization on CIFAR-10 using a 5 layer neural network as described in section \ref{5layers}.
    The red numbers in the figures indicate
    the percentages of ``corrupted labels'' used in pretraining. The
    green stars in the figures indicate the precise locations of the
    points. \textbf{Top left:} Training and test losses of
    unnormalized networks: there is no apparent relationship.
    \textbf{Top right:} the product of L2 norms from all layers of the
    network. We observe a positive correlation between the norm of the
    weights and the testing loss .  \textbf{Bottom:} under layerwise
    normalization of the weights (using the Frobenius norm), the
    classification error does not change (bottom right) while the
    cross-entropy loss changes (bottom left). There is a surprisingly good
    linear relationship between training and testing losses, implying
    tight generalization.  }
  \label{RLNLcifar10}  
\end{figure}

\begin{figure}[H]\centering
	\includegraphics[width=1.0\textwidth]{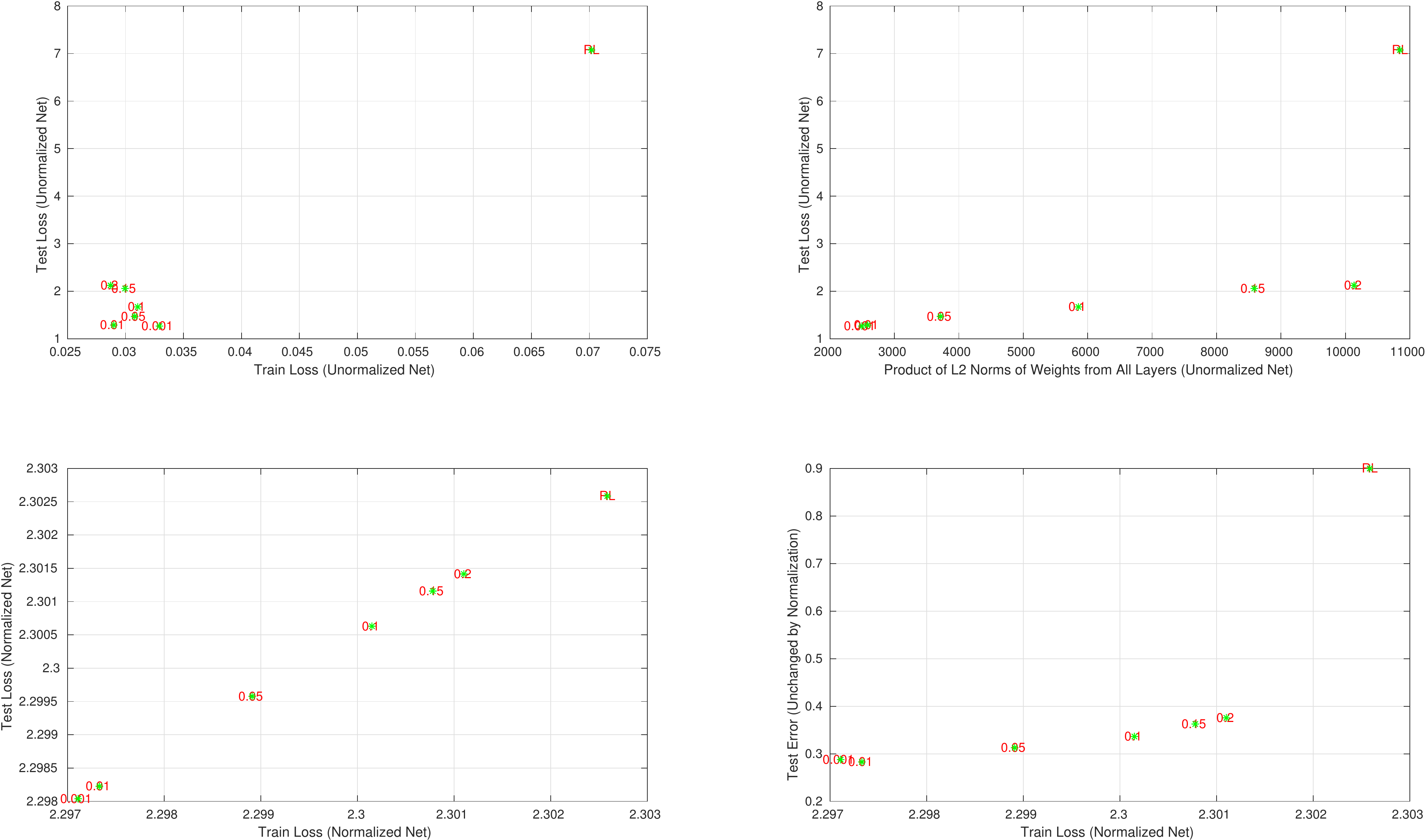}   
	\caption{\it Same as Figure \ref{RLNLcifar10} but using
          different standard deviations for initialization of the
          weights instead of ``random pretraining''. The red numbers
          in the figures indicate the standard deviations used in
          initializing weights. The ``RL'' point (initialized with
          standard deviation 0.05) refers to training and testing on
          completely random labels.  }
	\label{SDcifar10} 
\end{figure}

Our arguments so far imply that among two unnormalized minimizers of
the exponential loss that achieve the same given small loss
$L=\epsilon$, the minimizer with higher product of the norms
$\rho_1, \cdots, \rho_K$ has the higher capacity  and thus the
highest expected loss. Experiments support this claim, see Figure
\ref{product_frobenius}. Notice the linear relationship of test loss with
increasing capacity on the top right panels of Figure
\ref{RLNLcifar10}, \ref{SDcifar10}, \ref{RLNLcifar100},
\ref{SDcifar100}.

\begin{figure}[H]\centering
    \includegraphics[width=1.0\textwidth]{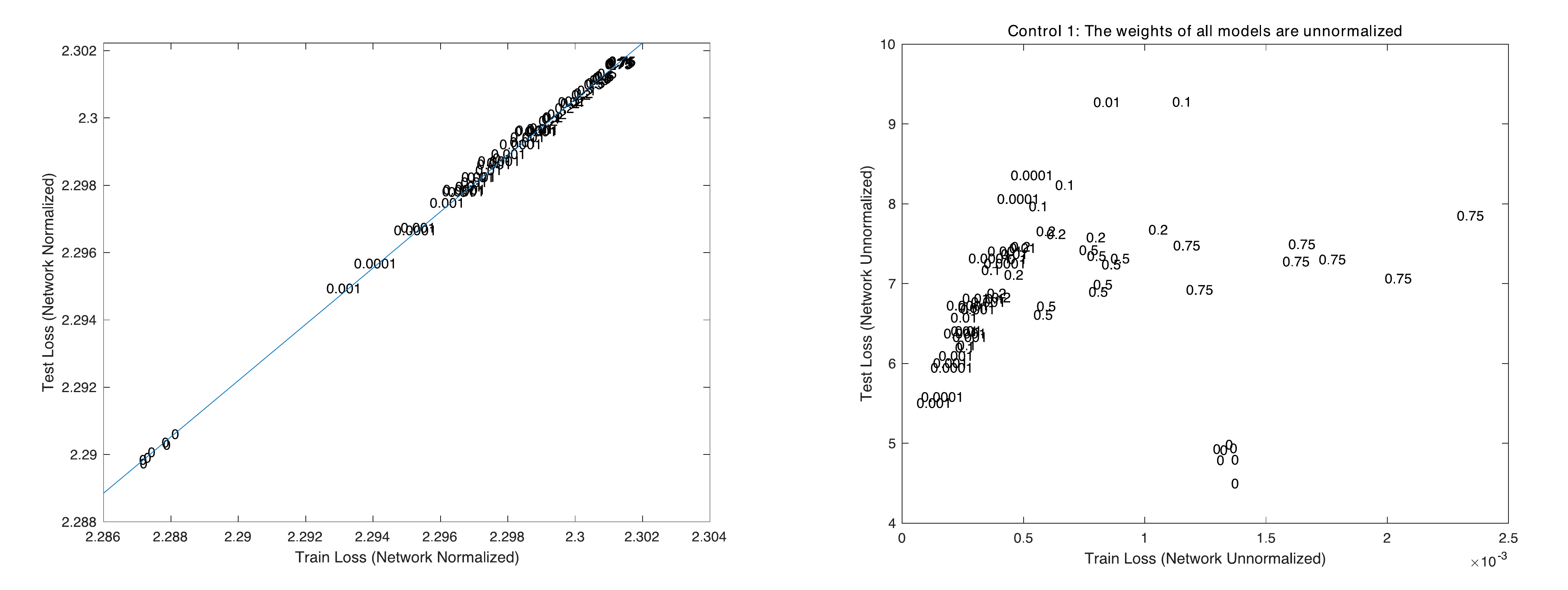}
    \caption{\it Left: test loss vs training loss with all networks
      normalized layerwise by the Frobenius norm.  Right: test loss vs
      training loss with all unnormalized networks.  The model was a 3
      layer neural network described in section \ref{3layers} and was
      trained with 50K examples on CIFAR10.  In this experiments the
      networks converged (and had zero train error) but not to the same
      loss.  All networks were trained for $300$ epochs.  The losses
      range approximately from $1.5 \times 10^{-4}$ to
      $2.5 \times 10^{-3}$.  The numbers in the figure indicate the
      amount of corruption of random labels used during pretraining.
      The slope and intercept of the line of best fit are $0.836$ and
      $0.377$ respectively.  The ordinary and adjusted $R^2$ values
      are both $0.9998$ while the root mean square (RMSE) is
      $4.7651 \times 10^{-5}$.  }
    \label{not_same_train_loss}
\end{figure} 

\section{Discussion}
\label{Generalization}

Our results support the theoretical arguments that an appropriate
measure of complexity for a deep network should be based on a
product norm\cite{AntBartlett2002},
\cite{DBLP:journals/corr/BartlettFT17},
\cite{DBLP:journals/corr/abs-1711-01530}. The most relevant recent
results are by \cite{DBLP:journals/corr/abs-1712-06541} where the
generalization bounds depend on the products of layerwise norms
without the  dependence on the number of layers present in previous
bounds. It is also interesting to notice that layerwise Frobenius norm
equal to $1$ as in our normalized networks avoids the implicit
dependence of the network complexity on the number of layers. The results from
\cite{Theory_IIIb}, which triggered this paper, imply products of
layerwise Frobenius norms.  As we discussed earlier, all $L_p$ norms
are equivalent in our setup.

\begin{figure}[H]\centering   
	\includegraphics[width=1.0\textwidth]{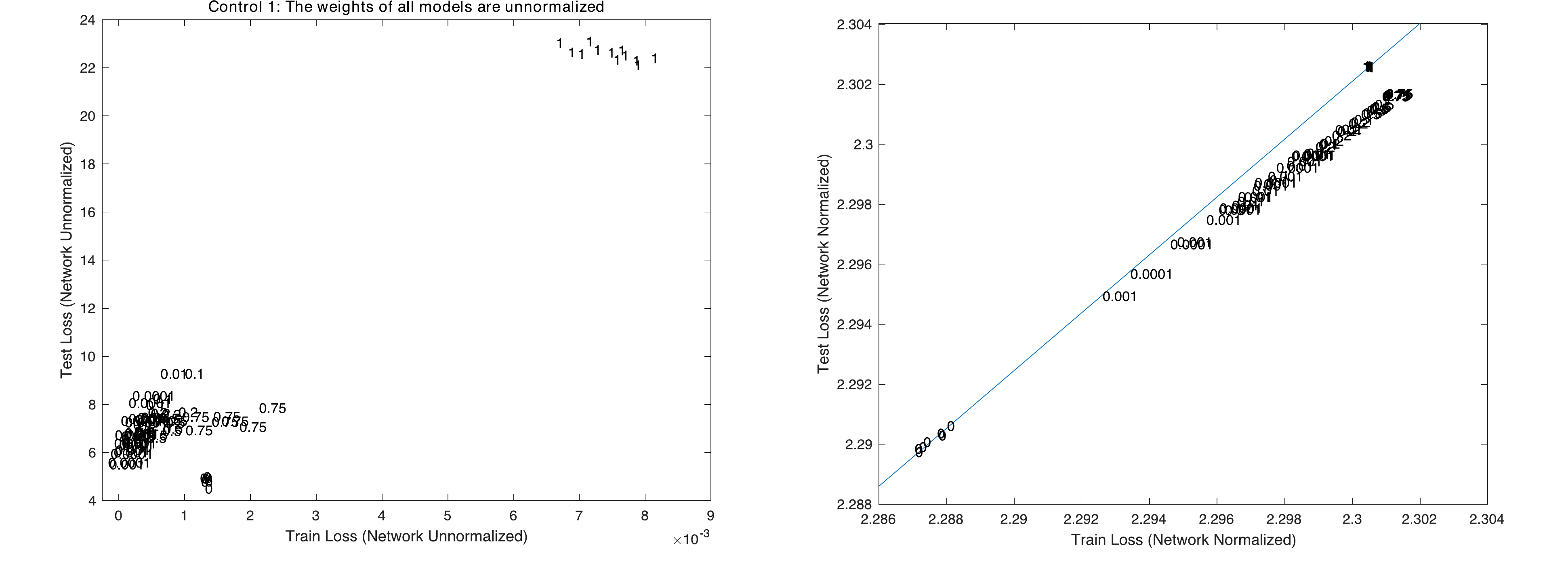}
	\caption{\it The left part of the figure shows the test
          cross-entropy loss plotted against the training loss for the
          normal, unnormalized network. Notice that all points have
          {\bf zero classification error at training}. The RL points
          have zero classification error at training and {\bf chance
            test error}. Once the trained networks are normalized
          layerwise, the training cross-entropy loss is a very good
          proxy for the test cross-entropy loss (right plot). The line
          -- regressed just on the $0$ and RL points -- with slope
          $0.9642$, offset $0.0844$, both ordinary and adjusted
          $R^2=0.9999$, root mean square (RMSE) was
          $6.9797 \times 10^{-5}$ seems to mirror well Equation
          \ref{bound}. }
	\label{upper_bounds_with_NL_vs_RL}
\end{figure}

The linear relation we found is quite surprising since it implies that
the classical generalization bound of Equation \ref{bound} is not only valid but is
{\it tight}: the {\it offset } in the right-hand side of the Equation
is quite small, as shown by Figures \ref{upper_bounds_with_NL_vs_RL}
and \ref{NL_vs_RL} (this also supported by Figures
\ref{not_same_train_loss}, \ref{L1} and \ref{only_NL_points}).  For
instance, the offset in Figure \ref{upper_bounds_with_NL_vs_RL} is
only $0.0844$.  This implies of course linearity in our plots of
training vs testing normalized cross-entropy loss. It shows that there
is indeed generalization -
defined as expected loss converging to training loss for large data
sets - in deep neural networks.  This result contradicts the claims in
the title of the recent and influential paper ``Understanding deep
learning requires rethinking generalization''
\cite{DBLP:journals/corr/ZhangBHRV16} though those claims {\it do apply} to
the unnormalized loss.

Though it is impossible to claim that better expected cross-entropy
loss implies better test error (because the latter is only an upper
bound to the classification error), our empirical results in various
of the figures show an approximatevely monotonic relation between
normalized test (or training) cross-entropy loss and test classification error with
roughly the shape of the $\psi$ transform of the logistic loss
function\cite{Bartlett03convexity}. Notice, in particular, that the normalized
cross-entropy loss in training for the randomly labeled set is close to
$\log 10$ which is the cross-entropy loss for chance 
performance found at test.

Though more experiments are necessary the linear relationship we found
seems to hold in a robust way across different types of networks,
different data sets and different initializations.

Our results, which are mostly relevant for theory, yield a
recommendation for practitioners: it is better to monitor during training the
empirical cross-entropy loss of the normalized network instead of the
unnormalized cross-entropy loss. The former is the one that matters in terms
of stopping time and test performance (see Figures \ref{L1} and
\ref{testlossvstrainloss}).

More significantly for the theory of Deep Learning, the observations
of this paper clearly demand a critical discussion of several commonly
held ideas related to generalization such as dropout, SGD and flat
minima.

\subsubsection*{Acknowledgments}
We thank Lorenzo Rosasco, Yuan Yao, Misha Belkin, Youssef Mroueh,
Amnon Sashua and especially Sasha Rakhlin for illuminating
discussions. Xavier Boix suggested the idea of random pretraining to
obtain get different test performances. NSF funding provided by CBMM.

	\small
\newpage
\bibliographystyle{plain}   

\bibliography{Boolean} \normalsize

\begin{center} 
  {\bf APPENDIX}
\end{center}	

\begin{appendices}

\section{Results on MNIST}

\begin{figure}[H]\centering   
	\includegraphics[width=1.0\textwidth]{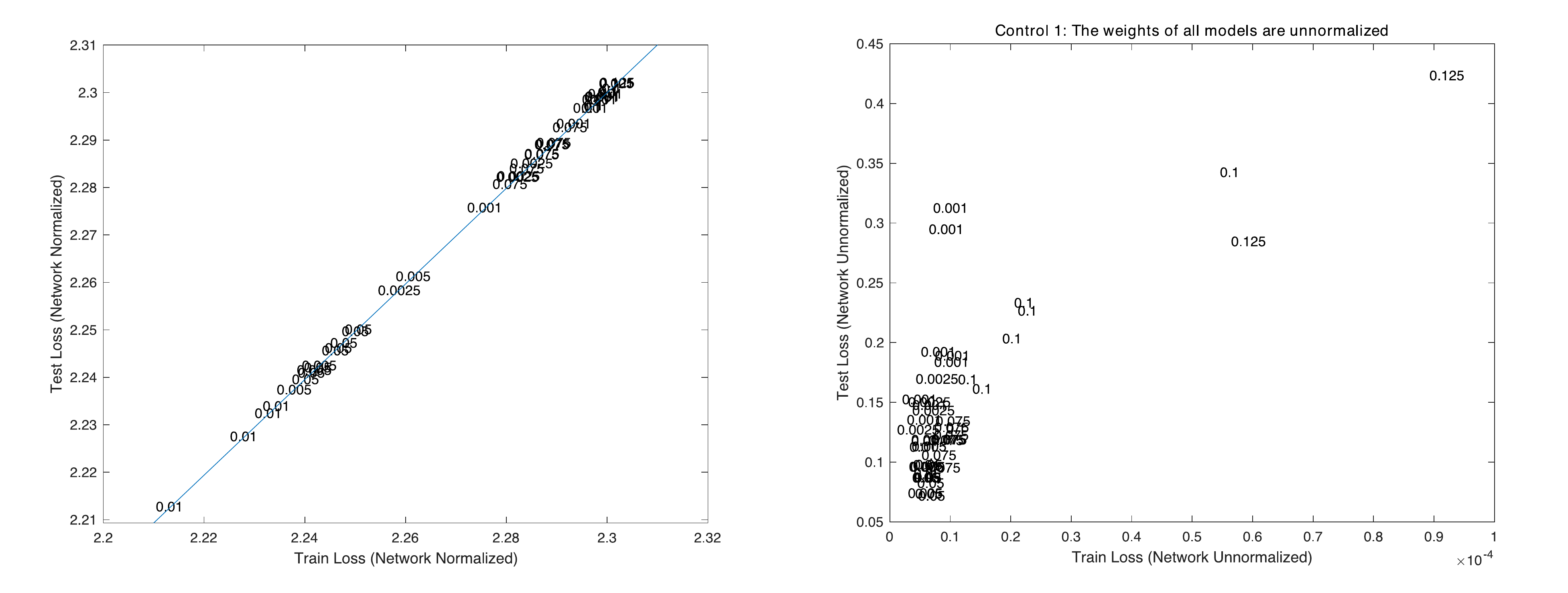}
	\caption{\it 
    The figure shows the cross-entropy loss on the test set vs the training loss for networks normalized layerwise in terms of the Frobenius norm.
    The model was a 3 layer neural network described in section \ref{3layers_34_units} and was trained with 50K examples on MNIST.
    All networks were trained for $800$ epochs. 
    In this experiments the networks converged (and had zero train error) but not to the same loss.
    The slope and intercept of the line of best fit are $1.0075$ and $-0.0174$ respectively. 
    The ordinary and adjusted $R^2$ values are both $1.0000$ while the 
    root mean square (RMSE) was $9.1093 \times 10^{-4}$.
    The makers indicate the size of the standard deviation of the normal used for initialization.
     }
	\label{mnist_nl}
\end{figure}

\begin{figure}[H]\centering   
	\includegraphics[width=1.0\textwidth]{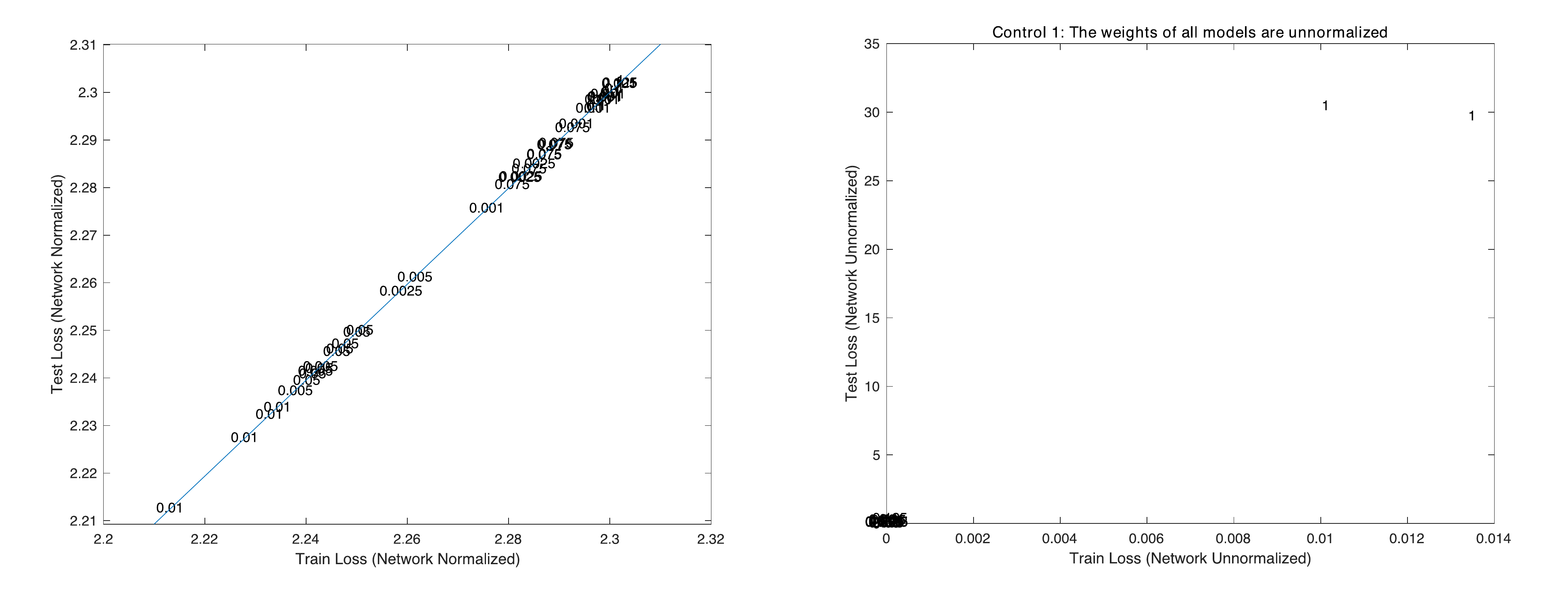}
	\caption{\it 
    The figure shows the cross-entropy loss on the test set vs the training loss for networks normalized layerwise in terms of the Frobenius norm.
    The model was a 3 layer neural network described in section \ref{3layers} and was trained with 50K examples on CIFAR10.
    All networks were trained for $800$ epochs. 
    In this experiments the networks converged (and had zero train error) but not to the same loss.
    The slope and intercept of the line of best fit are $1.0083$ and $-0.0191$ respectively. 
    The ordinary and adjusted $R^2$ values are both $1.0000$ while the 
    root mean square (RMSE) was $9.1093 \times 10^{-5}$.
    The points labeled $1$ were trained on
    random labels; the training loss was estimated on the same
    randomly labeled data set.  The points marked with values less than $1$ were
    only trained on natural labels and those makers indicate the size of the standard deviation of the normal used for initialization.
     }
	\label{mnist_rl_vs_nl}
\end{figure}

This section shows figures replicating the main results on the MNIST data set. 
Figures \ref{mnist_nl} and \ref{mnist_rl_vs_nl} show that the linear relationship holds after normalization on the MNIST data set.
Figure \ref{mnist_rl_vs_nl} shows the linear relationship holds after adding the point trained only on random labels.

\section{Results on CIFAR-100}

\begin{figure}[H]\centering
	\includegraphics[width=0.75\textwidth]{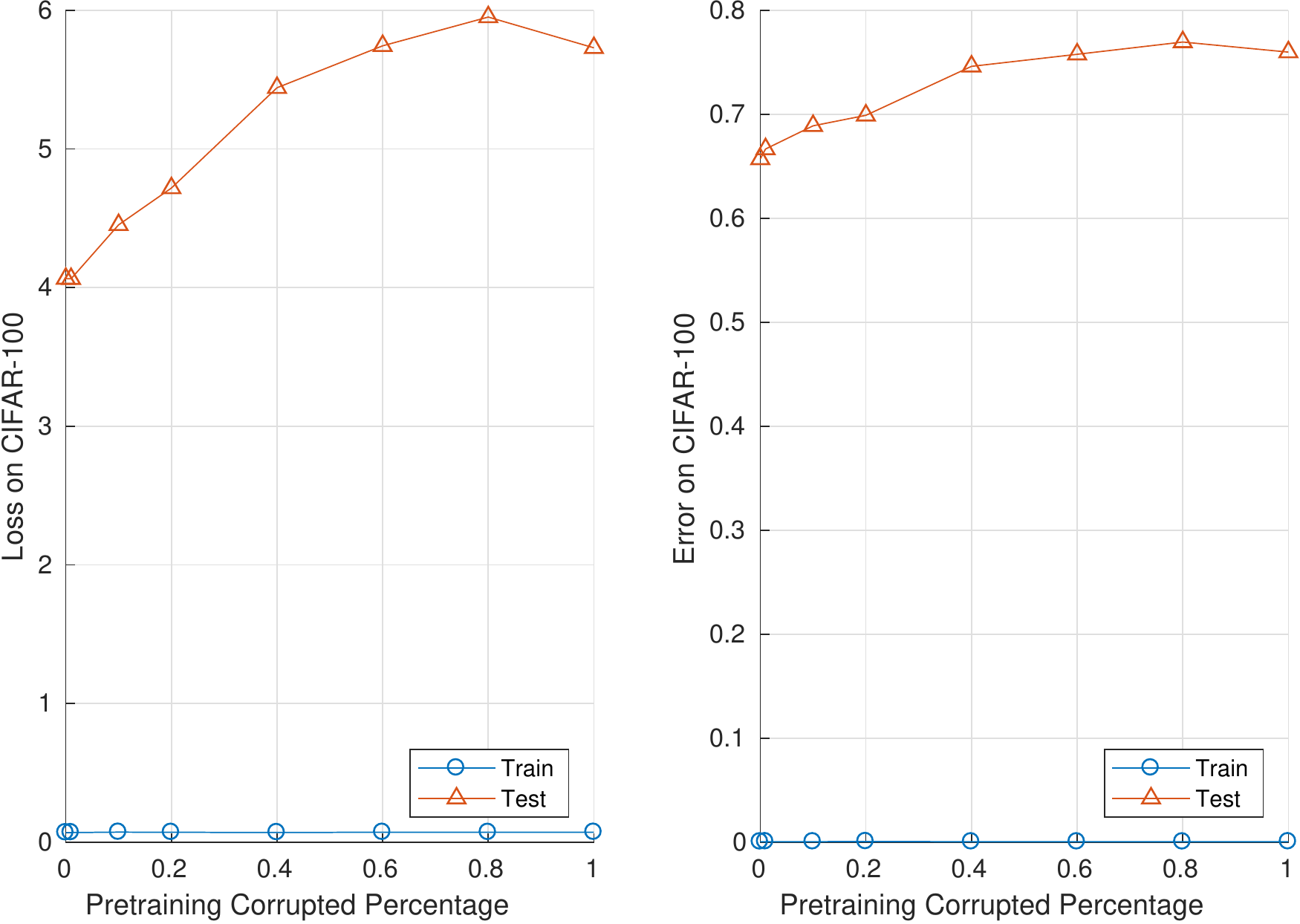}
	\caption{\it Same as Figure \ref{cifar10_RLNL_performance}, but on CIFAR-100. }    
	\label{cifar100_RLNL_performance}    
\end{figure}

\begin{figure}[H]\centering
	\includegraphics[width=0.75\textwidth]{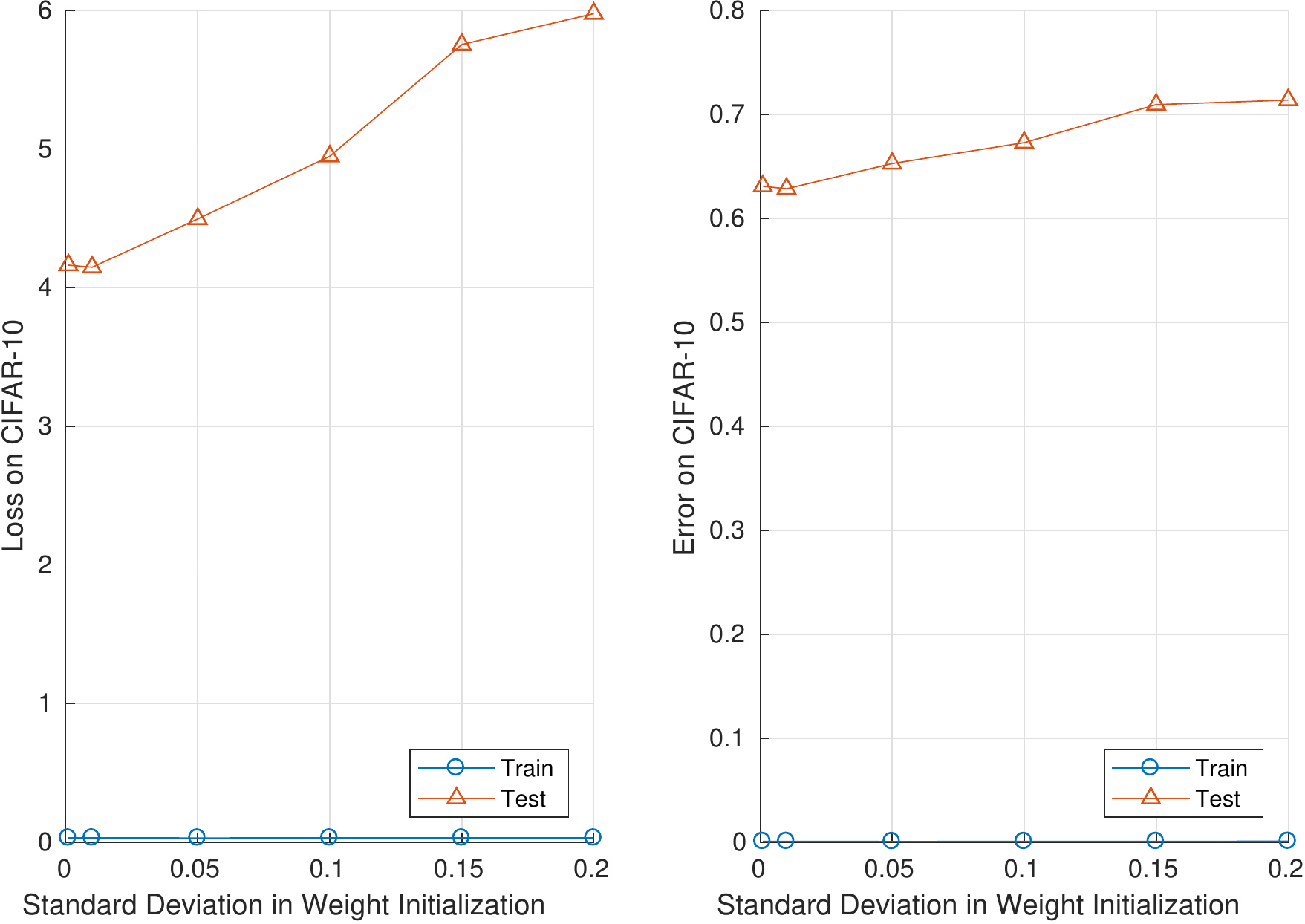} 
	\caption{ \it Same as Figure \ref{cifar10_DMag_performance}, but on CIFAR-100.}   
	\label{cifar100_DMag_performance}            
\end{figure}

This section shows figures replicating the main results on CIFAR-100.
Figure \ref{cifar100_RLNL_performance} shows how different test
performance can be obtained with pretraining on random labels while
Figure \ref{cifar100_DMag_performance} shows that different increasing
initializations are also effective.

\begin{figure}[H]\centering 
	\includegraphics[width=1.0\textwidth]{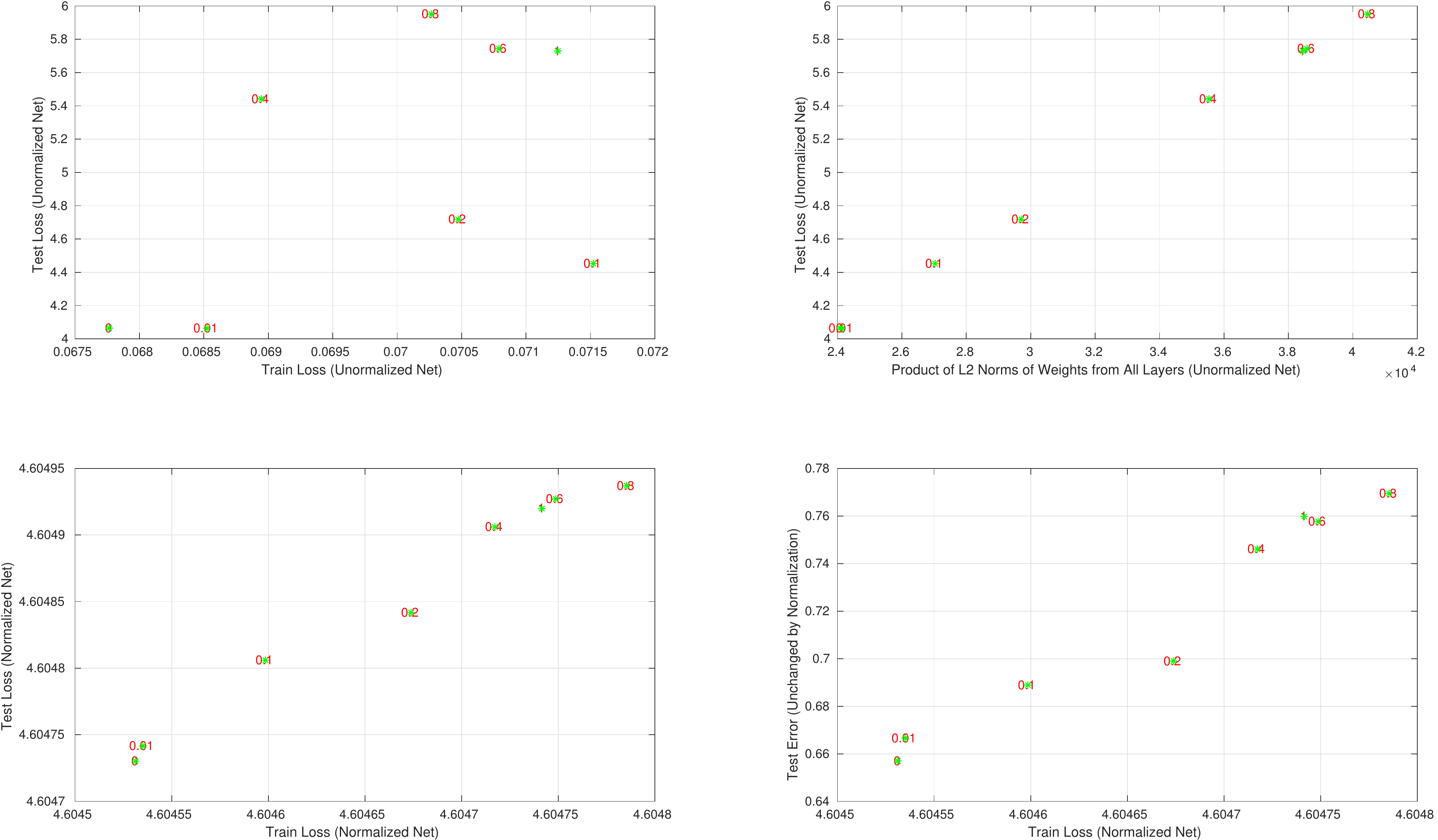} 
	\caption{ \it Same as Figure \ref{RLNLcifar10} but on CIFAR-100  }     
	\label{RLNLcifar100} 
\end{figure}

\begin{figure}[H]\centering      
	\includegraphics[width=1.0\textwidth]{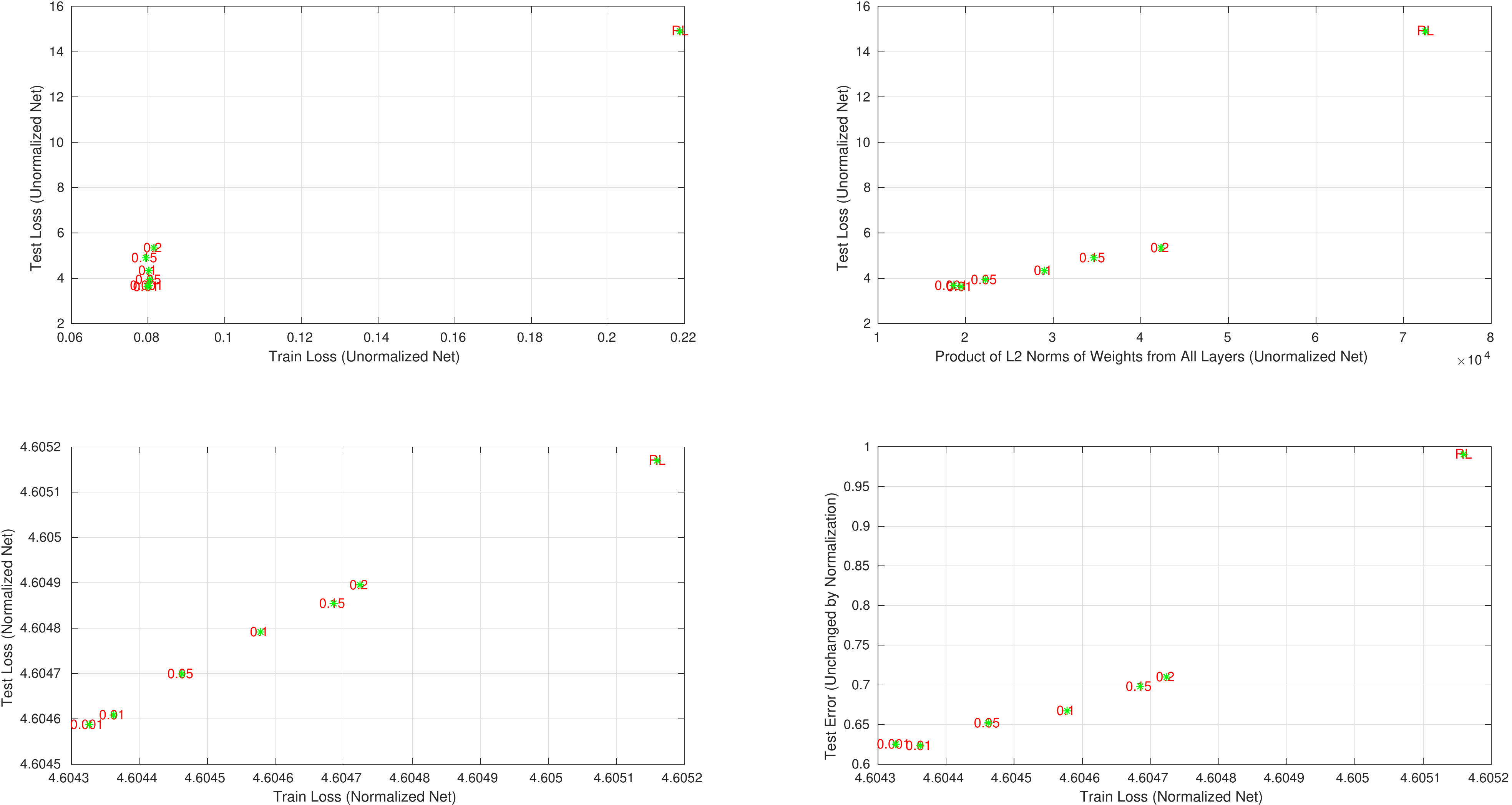}
	\caption{\it Same as Figure \ref{SDcifar10} but on CIFAR-100.  }            
	\label{SDcifar100}          
\end{figure}

More importantly, Figures \ref{RLNLcifar100} and \ref{SDcifar100} show
that the linear relationship holds after normalization, regardless of
whether the training was done with pretraining on random labels or
with large initialization.


\section{$L_p$ norms for normalization}  

\begin{figure}[H]\centering
	\includegraphics[width=1.0\textwidth]{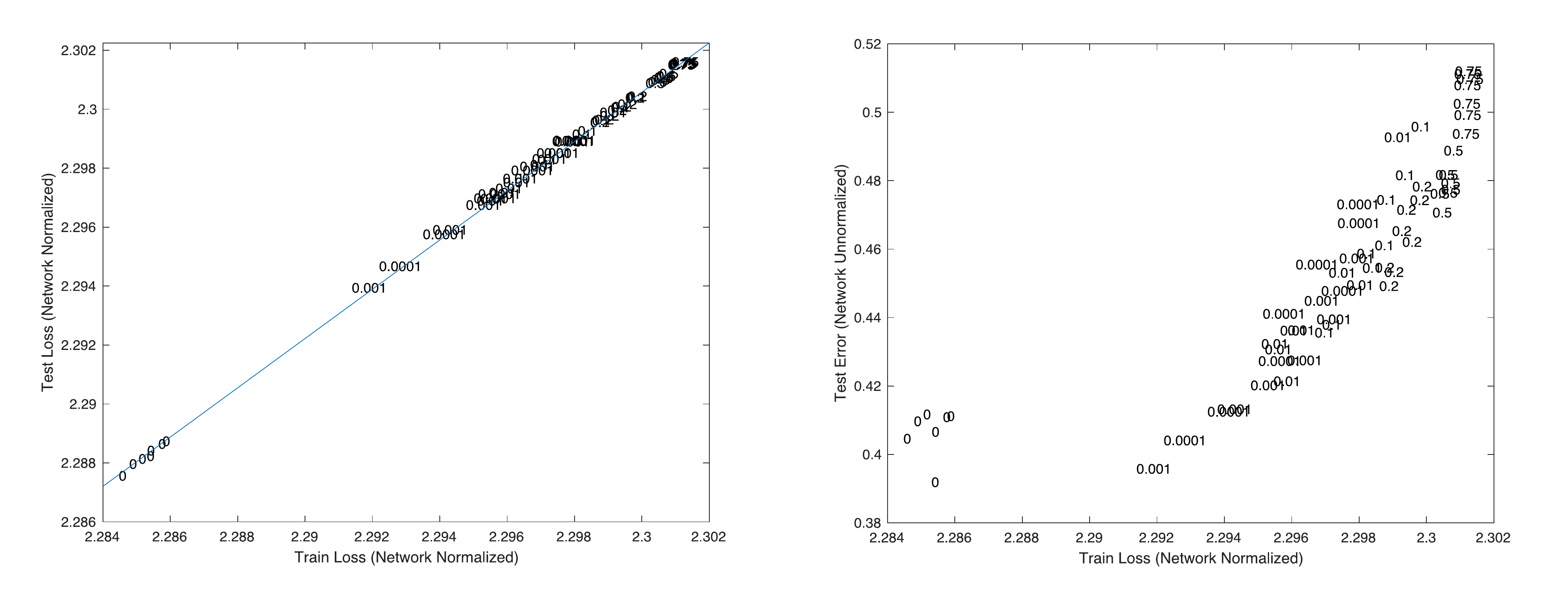}
	\caption{\it Test loss/error vs training loss with all
          networks normalized layerwise by the L1 norm (divided by
          $100$ to avoid numerical issues because the L1 norms are
          here very large).  The model was a 3 layer neural network
          described in section \ref{3layers} and was trained with 50K
          examples on CIFAR10.  The networks were normalized after
          training each up to  epoch $300$ and thus different points on the plot 
          correspond to different training losses. Figure
          \ref{not_same_train_loss} shows that  the normalized network
          does not depend on the value of the
          training loss before normalization.  The slope and
          intercept of the line of best fit are $0.8358$ and $0.3783$
          respectively.  The ordinary and adjusted $R^2$ values are
          both $0.9998$ while the root mean square (RMSE) was
          $5.6567 \times 10^{-5}$.  The numbers in the figure indicate
          the amount of corruption of random labels used during the
          pretraining.  }
	\label{L1}
\end{figure}

This section verifies that the effectiveness of layerwise
normalization does not depend on which $L_p$ norm is used.  Figure
\ref{L1} provides evidence for this.  The constants underlying the
equivalence of different norms depend on the dimensionality of the
vector spaces.

\section{Additional Evidence for a Linear Relationship}

This section provides additional experiments  showing a linear
relationship between the test loss and the training loss after
layerwise normalization.

\begin{figure}[H]\centering
	\includegraphics[width=1.0\textwidth]{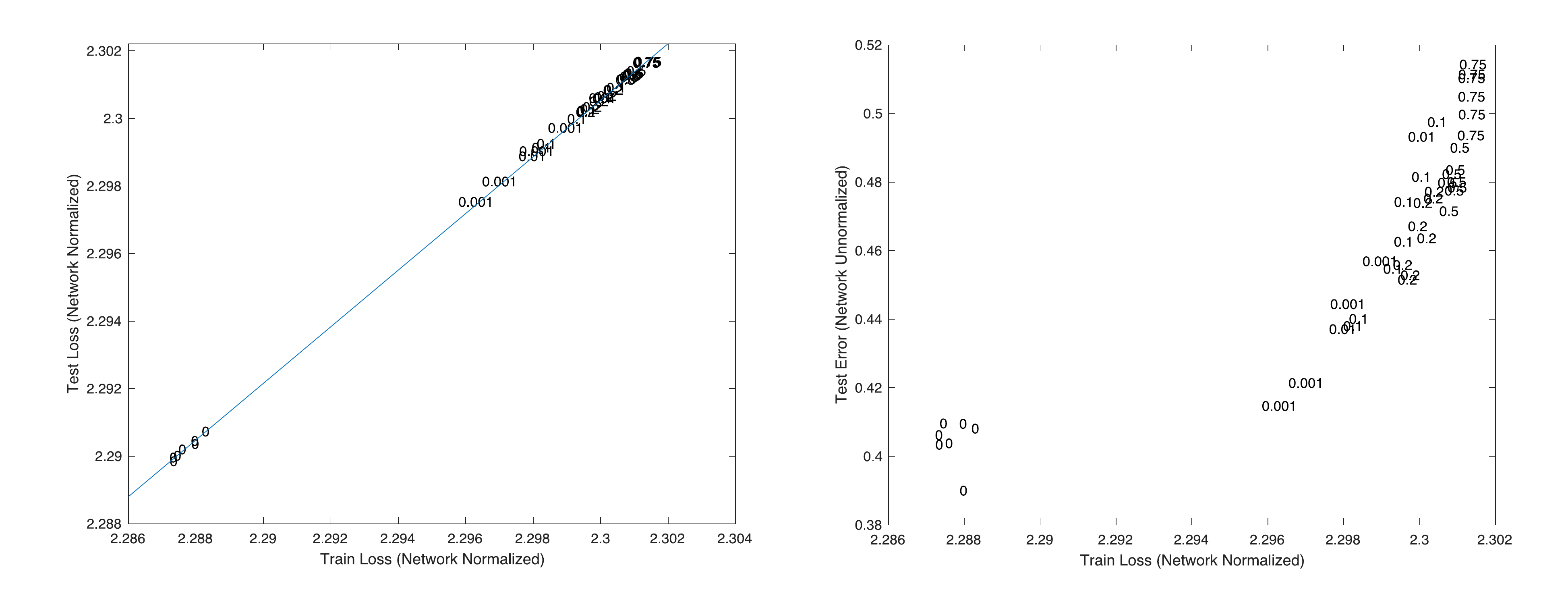}
	\caption{\it Test loss/error vs training loss with all
          networks normalized layerwise by the Frobenius norm of the
          weights. The model was a 3 layer neural network described in
          section \ref{3layers} and was trained with 50K examples on
          CIFAR10.  The models were obtained by pretraining on random
          labels and then by fine tuning on natural labels. SGD
          without batch normalization was run on all networks in this
          plot until each reached approximately $0.0044 \pm 0.0001$
          cross-entropy loss on the training data.  The numbers in the
          figure indicate the amount of corruption of the random
          labels used in pretraining.}
	\label{testlossvstrainloss}
\end{figure}

Figure \ref{testlossvstrainloss} shows that the linear relationship
holds if all models are stopped at approximately the same train loss.

\begin{figure}[H]\centering
    \includegraphics[width=1.0\textwidth]{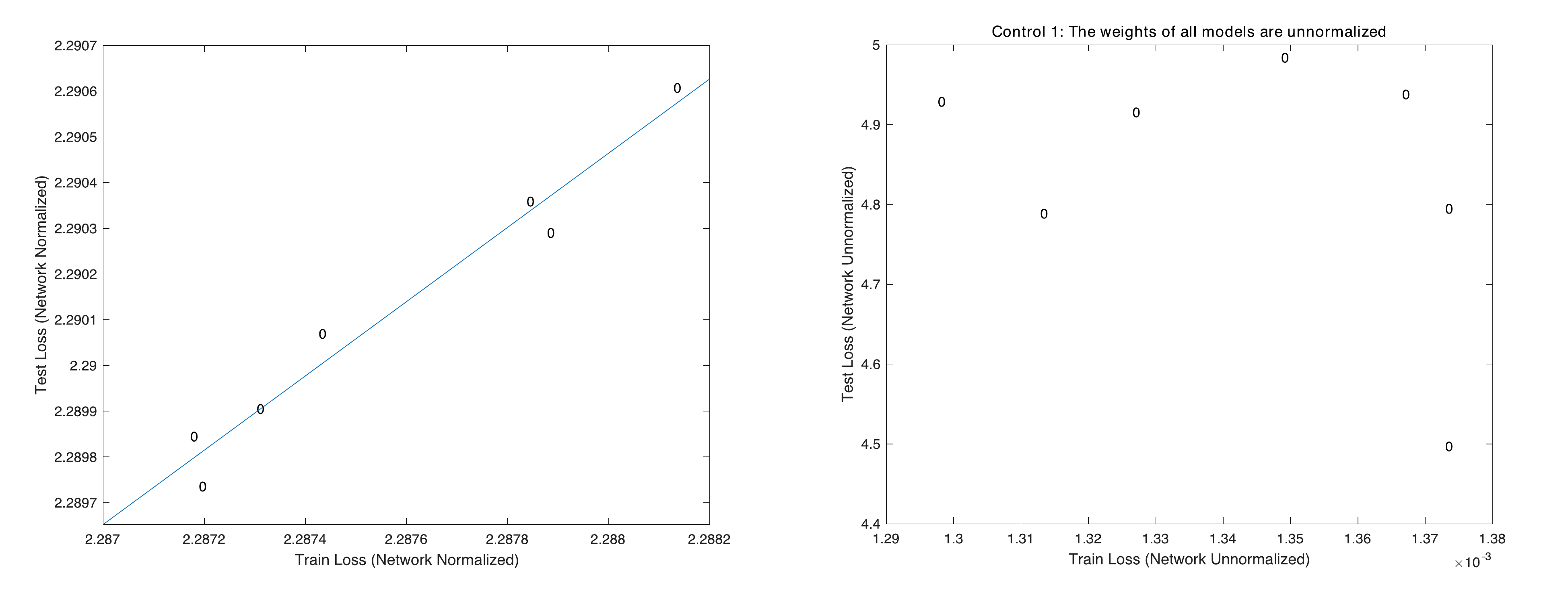}
    \caption{\it Left: test loss vs training loss for networks
      normalized layerwise by the Frobenius norm.  Right: test loss
      vs training loss for unnormalized networks.  The model was a 3
      layer neural network described in section \ref{3layers} and was
      trained with 50K examples on CIFAR10.  In this experiments the
      networks converged (and had zero train error) but not to the same
      loss.  The networks were trained for $300$ epochs.  The cross
      entropy training losses range approximately from
      $1.29 \times 10^{-3}$ to $1.38 \times 10^{-3}$.  The $0$s in the
      figure indicate that there was no random pretraining.  The
      slope and intercept of the line of best fit are $0.8117$ and
      $0.4333$ respectively.  The ordinary $R^2$ is $0.9660$ and the
      adjusted $R^2$ value is $0.9592$ while the root mean square
      (RMSE) was $6.3624 \times 10^{-5}$.  }
    \label{only_NL_points} 
\end{figure}

Figure \ref{only_NL_points} is the same as Figure
\ref{upper_bounds_with_NL_vs_RL}, except that the randomly pretrained
networks were removed.  The important thing to note is that the {\it
  offset} is $0.0844$, which implies that the \ref{bound} is surprisingly
tight.

\begin{figure}[H]\centering   
	\includegraphics[width=1.0\textwidth]{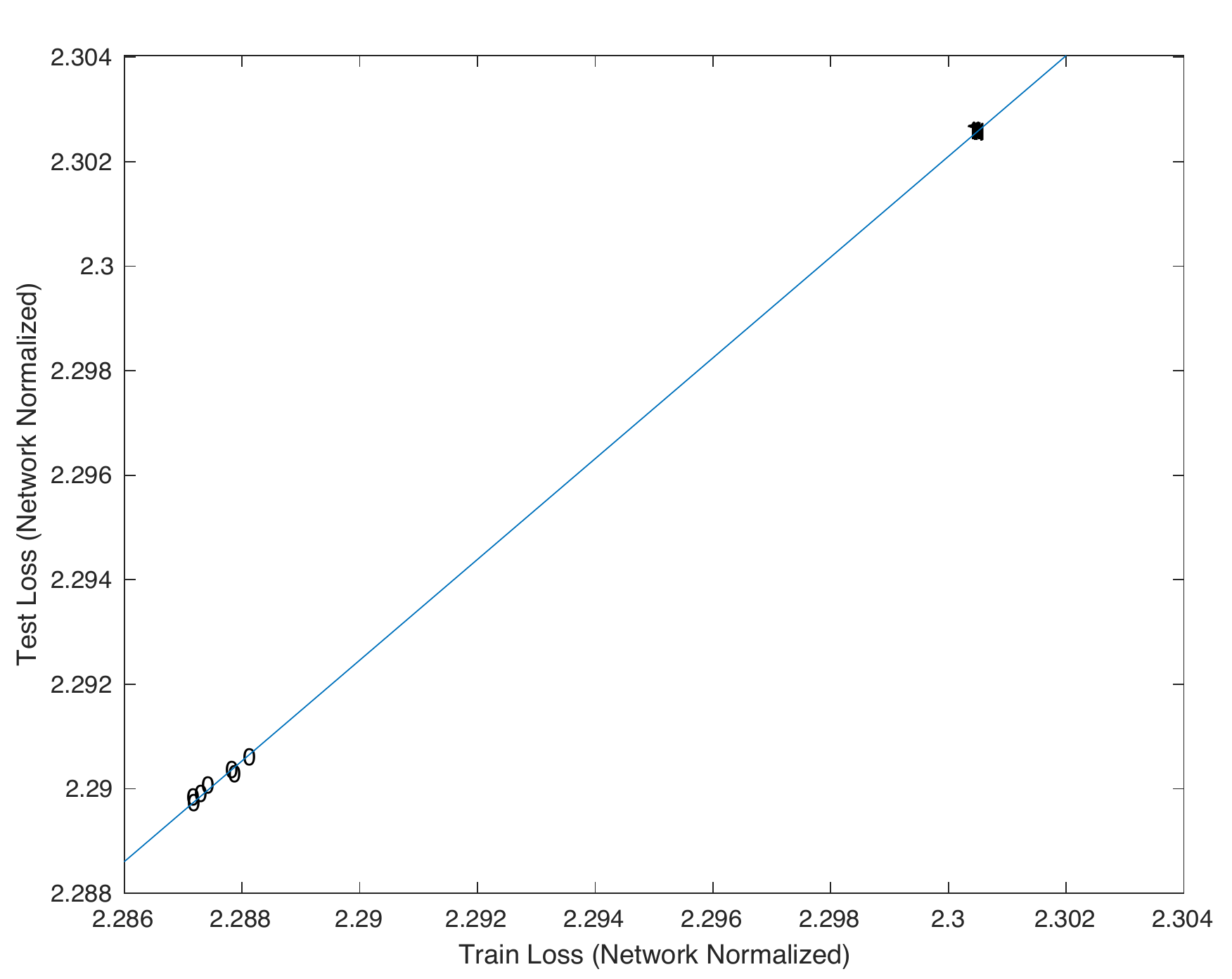}
	\caption{\it 
    The figure shows the cross-entropy loss on the test set vs the training loss for networks normalized layerwise in terms of the Frobenius norm.
    The model was a 3 layer neural network described in section \ref{3layers} and was trained with 50K examples on CIFAR10.
    All networks were trained for $300$ epochs. 
    In this experiments the networks converged (and had zero train error) but not to the same loss.
    The slope and intercept of the line of best fit are $0.9642$ and $0.0844$ respectively. 
    The ordinary and adjusted $R^2$ values are both $0.9999$ while the 
    root mean square (RMSE) was $6.9797 \times 10^{-5}$.
    The points labeled $1$ were trained on
    random labels; the training loss was estimated on the same
    randomly labeled data set.  The points marked with $0$ were
    only trained on natural labels.
     }
	\label{NL_vs_RL}
\end{figure}

\begin{figure}[H]\centering
	\includegraphics[width=1.0\textwidth]{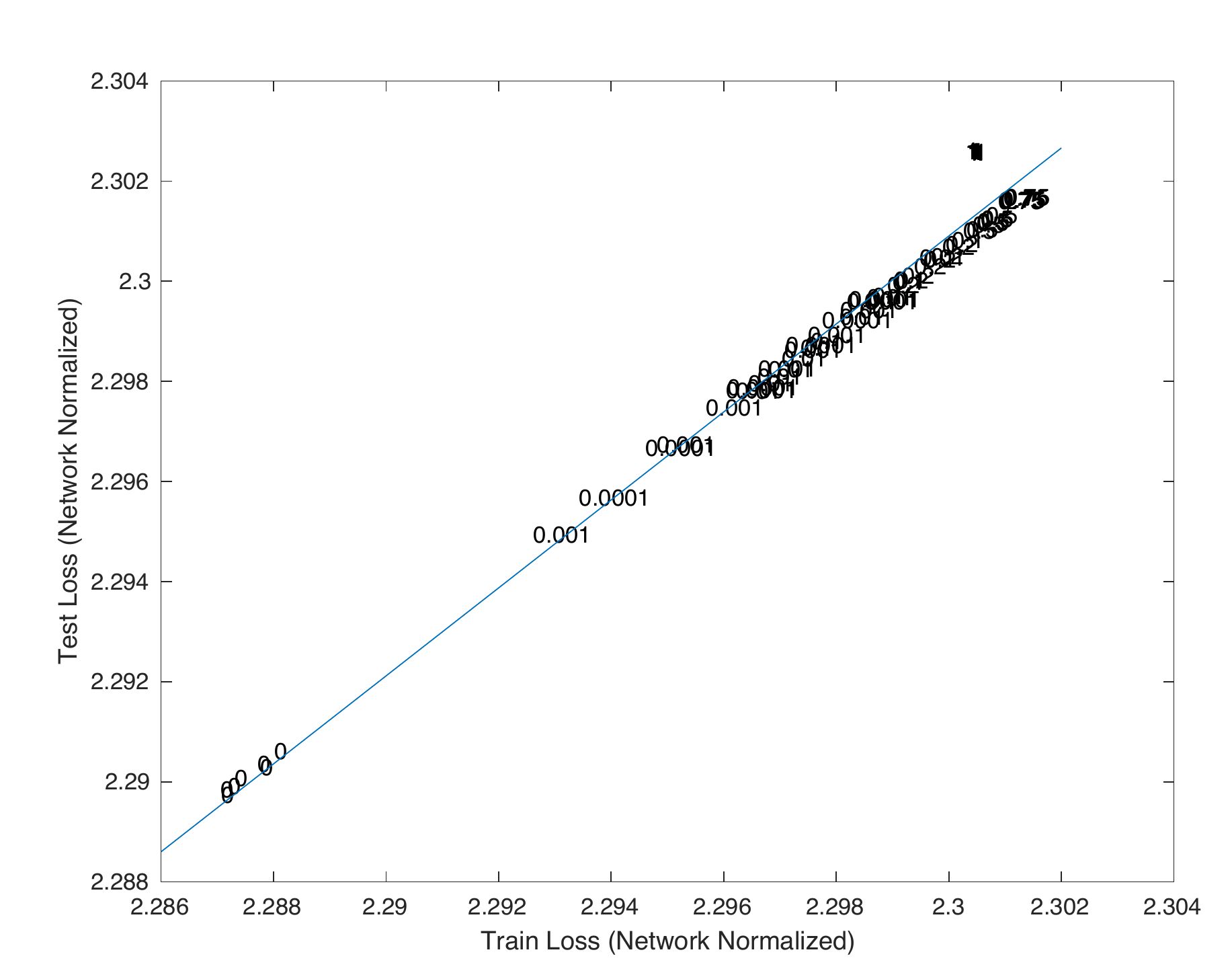}
	\caption{\it  The figure shows cross-entropy loss on the test set vs the training loss for networks normalized layerwise in terms of the Frobenius norm.
    The model was a 3 layer neural network described in section \ref{3layers} and was trained with 50K examples on CIFAR10.
    All networks were trained for $300$ epochs. 
    In this experiments the networks converged (and had zero train error) but not to the same loss.
    The slope and intercept of the line of best fit are $0.8789$ and $0.2795$ respectively. 
    The ordinary and adjusted $R^2$ values are both $0.9721$ while the root mean square (RMSE) was $5.8304 \times 10^{-4}$.
    See Figure \ref{NL_vs_RL} for
    other details.
     }
	\label{RL_point}
\end{figure}

\section{Higher Capacity leads to higher Test Error}

\begin{figure}[H]\centering
	\includegraphics[width=1.0\textwidth]{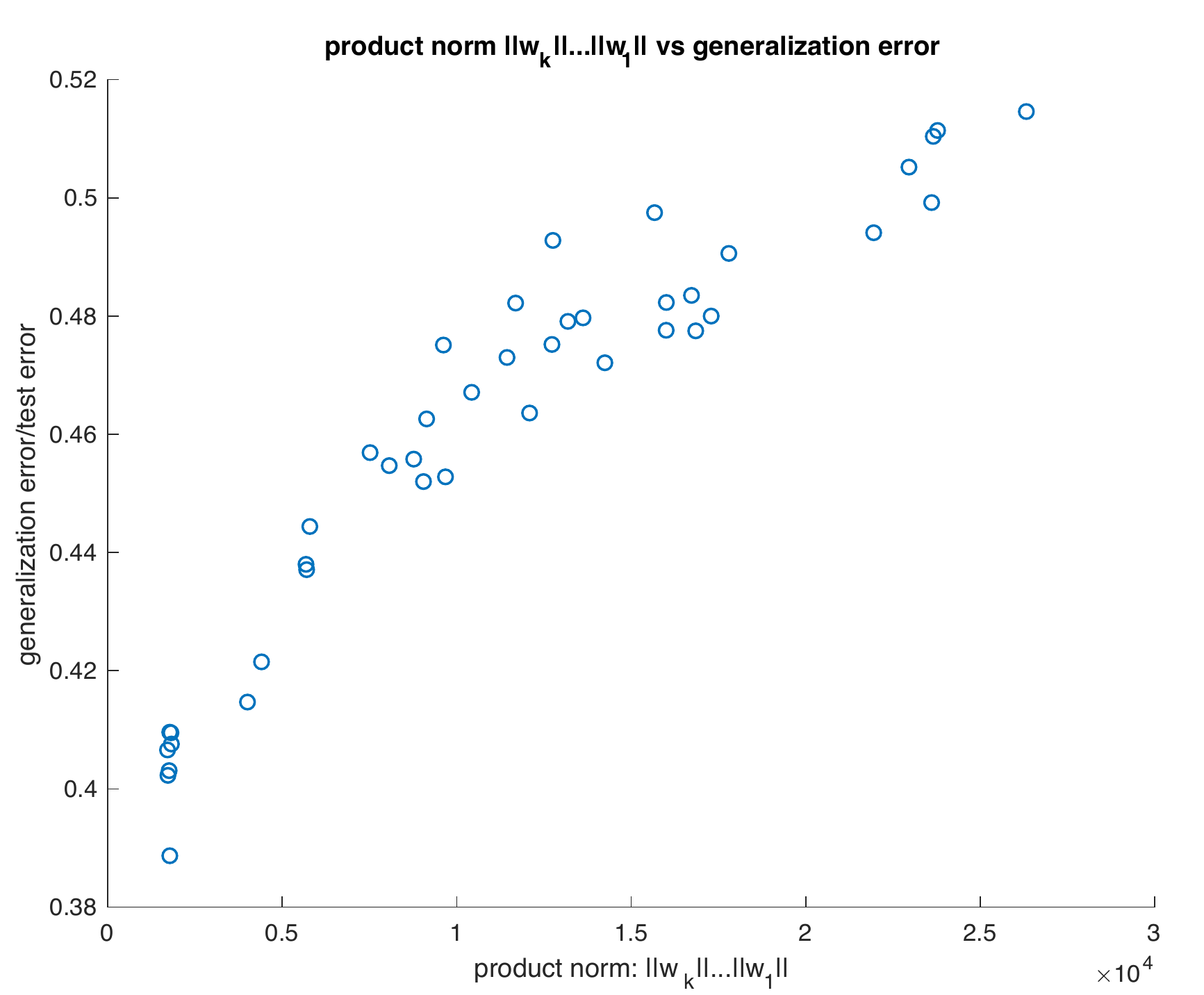}            
	\caption{\it 
	  Plot of test error vs the product of the Frobenius norms of the layers $\ \| W \|_{product} = \prod^L_{l=1} \| W_l \|$.
          The model was a 3 layer neural network described in section \ref{3layers} and  trained
          with 50K examples on CIFAR10. The models were obtained by
          pretraining on random labels and then fine tuning on
          natural labels. SGD without batch normalization
          was run on all networks in this plot until each reached
          approximately $0.0044 \pm 0.0001$ cross-entropy loss on the
          training data.  }
	\label{product_frobenius}
\end{figure}

Figure \ref{product_frobenius} shows that when the capacity of a network (as measured by the product norm of the layers)increases, so does the test error.

\section{Numerical values of normalized loss}

\begin{figure}[H]\centering
	\includegraphics[width=1.0\textwidth]{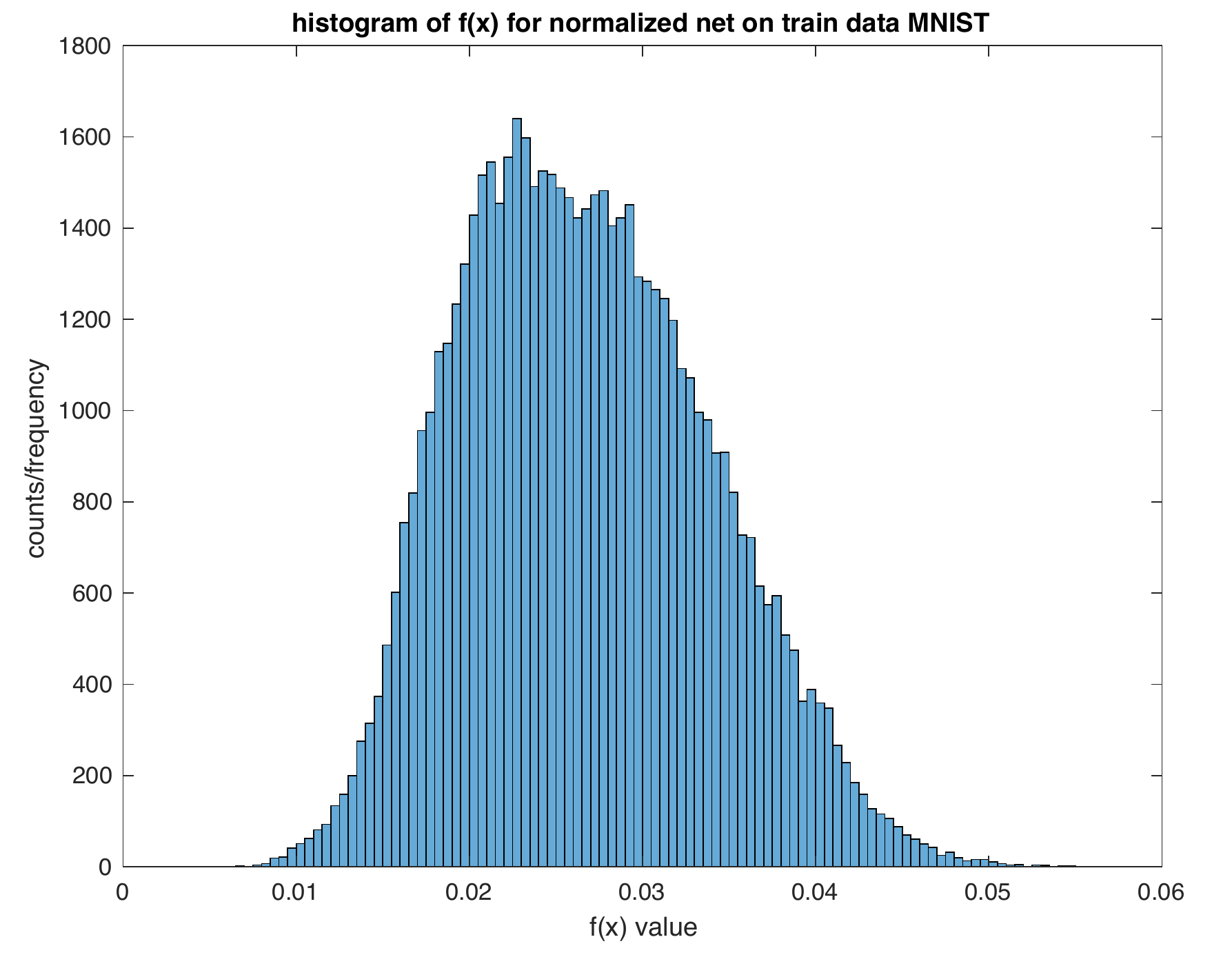}
	\caption{\it 
	  Histogram of the values of $f(x)$ for the most likely class of the layerwise normalized neural network over the 50K images of the MNIST training set.   
	  The average value $f(x)$ of the most likely class according to the 
	  normalized neural network is $0.026683$ with standard deviation $0.007144$. 
	  }
	\label{hist_train_mnist_norm}
\end{figure}

In this section we discuss very briefly some of the intriguing
properties that we observe after layerwise normalization of the
neural network.  Why are all the cross-entropy loss values close to chance
(e.g. $\ln 10 \approx 2.3$ for a $10$ class data set) in all the plots
showing the linear relationship?  This is of course because most of
the (correct) outputs of the normalized neural networks are close to
zero as shown by Figure \ref{hist_train_mnist_norm}. The reason for this 
is that we would roughly expect the norm of the network to be bounded by a norm
without the ReLU activations $|f(\tilde{W};x)|
\lessapprox |\tilde{W}| |x| = |x|$, and the data $x$ is usually
pre-processed to have mean 0 and a standard deviation of 1.  
In fact, for the MNIST experiments, the
average value $f(x) $ of the most likely class according to the
normalized neural network is $0.026683$ with a standard deviation
$0.007144$.  This means that significant differences directly
reflecting the predicted class of each point are between $0.019539$ and
$0.033827$. This in turn implies that the exponentials in the cross-entropy
loss are all very close to 1. 

\begin{figure}[H]\centering
	\includegraphics[width=1.0\textwidth]{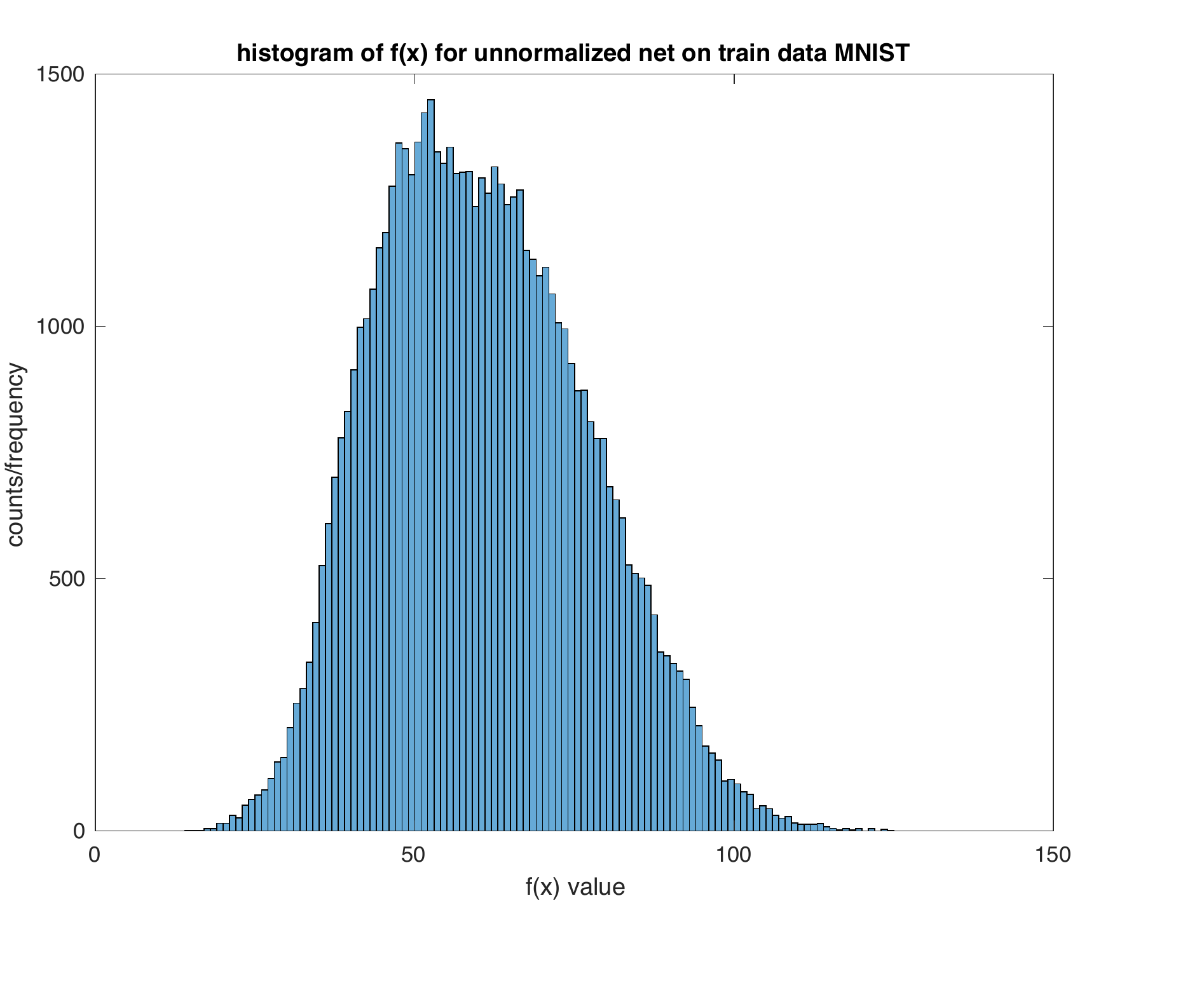} 
	\caption{\it 
	  Histogram of the values of $f(x)$ for the most likely class of the unnormalized neural network over the 50K images of the MNIST training set.
	  The average value $f(x)$ of the most likely class according to the 
	  unnormalized neural network is $60.564373$ with standard deviation $16.214078$.
	  }
	\label{hist_train_mnist_un}
\end{figure}

As a control experiment refer to Figure
\ref{hist_train_mnist_un} for the histogram of the final layer for the
most likely class for the unnormalized network.  Note that not only
are the differences between loss values far from machine precision
$\sim 10^{-7.5}$, but also that the output values of the normalized
neural networks are far from machine precision too as shown by Figure
\ref{hist_train_mnist_norm}. Notice that normalization does not affect
the classification loss. In summary, though the values of the points
in the figures differ slightly in terms of loss from each other, those
differences are highly significant and reproducible.

\section{Deep Neural Network Architecture }

\subsection{ Three layer network }
\subsubsection{Network with $24$ filters}
\label{3layers}

The model is a $3$-layer convolutional ReLU network with the first two
layers containing $24$ filters of size $5$ by $5$; the final layer is
fully connected; only the first layer has biases. There is no pooling.

The network is overparametrized: it has  $154,464$ parameters (compared to
$50,000$ training examples).

\subsubsection{Network with $34$ filters}
\label{3layers_34_units}

The model is the same $3$-layer convolutional ReLU network as in section \ref{3layers} except it had $34$ units.

The network was still overparametrized: it has  $165,784$ parameters (compared to
$50,000$ training examples).

\subsection{ Five layer network }
\label{5layers}

The model is a $5$-layer convolutional ReLU network with (with no
pooling).  It has in the five layers $32,64,64,128$ filters of size
$3$ by $3$; the final layer is fully connected; batch-normalization is
used during training.

The network is overparametrized with about $188,810$ parameters
(compared to $50,000$ training examples).

\section{Intuition: Shallow Linear Network} 
Consider a shallow linear network, trained once with GD on a set of $N$  
data $ \{ x^a_i,y^a_i \}^N_{i=1}$ and separately on a different set of $N$ data $ \{ x^b_i,y^b_i \}^N_{i=1}$. Assume
that in both cases the problem is linearly separable and thus zero
error is achieved on the training set. The question is
which of the two solutions will generalize better. The natural
approach is to look at the $L_2$ norm of the solutions:
$\sum_n (y_n -w^Tx_n)^2$

\begin{equation}
 f(w^a,x^a)= \langle w^a, x^a \rangle \quad \quad  f(w^b,x^b)= \langle w^b, x^b \rangle
\end{equation}
 
In both cases GD converges to zero loss with the minimum norm, maximum
margin solution for $w$. Thus the solution with the larger margin (and
the smaller norm) should have a lower expected error.  In fact,
generalization bounds (\cite{kakade2009complexity}) appropriate for
this case depend on the product of the norm $\|w\|$ and of a bound on
the norm of the data.

\end{appendices}

\end{document}